\documentclass[a4paper,fleng]{article}

\usepackage[left=1.5cm, right=1.5cm, top=1.5cm, bottom=1.5cm]{geometry}

\usepackage[blocks]{authblk}
\usepackage[colorlinks=true, allcolors=blue]{hyperref}
\usepackage{float}
\usepackage{graphicx}
\usepackage{amsmath}
\usepackage{array, multirow}
\usepackage{tabularx}
\usepackage{adjustbox}
\usepackage{booktabs}
\usepackage{enumitem}
\usepackage{algorithm}
\usepackage{algpseudocode}
\usepackage{subfig}
\usepackage{hyperref}
\usepackage{xcolor}
\usepackage{appendix}
\usepackage{soul}
\usepackage{boldline}
\usepackage{setspace} 
\title{Effective Self Attention-Based Deep Learning Model with Evolutionary Grid Search for Robust Wave Farm Energy Forecasting}

\author[1,7]{Amin Abdollahi Dehkordi}

\affil[1]{Department of Computer Engineering, Najaf Abad Azad University \\ \texttt {amin.abdollahi.dehkordi@azad.edu.ir}}

\author[2,3]{Mehdi Neshat}

\affil[2]{Faculty of Engineering and Information Technology, University of Technology Sydney, Ultimo \\ \texttt{mehdi.neshat@uts.edu.au}}
\author[3]{Nataliia Y. Sergiienko}

\affil[3]{School of Electrical and Mechanical Engineering, University of Adelaide \\ \texttt{nataliia.sergiienko@adelaide.edu.au}, \texttt{Zahra.Ghasemi@adelaide.edu.au} , \texttt{Lei.Chen@adelaide.edu.au}}

\author[3]{Zahra Ghasemi}

\author[3]{Lei Chen}

\author[4]{John Boland}

\affil[4]{Industrial AI Research Centre, UniSA STEM, University of South Australia \\ \texttt{John.Boland@unisa.edu.au}}

\author[5]{Hamid Moradkhani}

\affil[5]{Center for Complex Hydrosystems Research (CCHR), The University of Alabama, Tuscaloosa, USA \\ \texttt{hmoradkhani@ua.edu} }
    
\author[2,6] {Amir H. Gandomi \\ \thanks{Corresponding author: \texttt {gandomi@uts.edu.au}}}

\affil[6]{University Research and Innovation Center (EKIK), Obuda University}
    
\affil[7]{School of Systems \& Computing, The University of New South Wales \\ \texttt{a.abdollahidehkordi@unsw.edu.au}}
\begin{document}
\maketitle
\let\WriteBookmarks\relax
\def\floatpagepagefraction{1}
\def\textpagefraction{.001}

\begin{abstract}
Achieving carbon neutrality and mitigating greenhouse gas emissions represent critical global imperatives, as emphasized by the United Nations Sustainable Development Goal (SDG) \#13. In this context, wave energy emerges as an up-and-coming renewable resource, theoretically capable of producing approximately 30,000 terawatt-hours (TWh) of clean electricity annually—exceeding current global electricity demand. Despite its vast potential, the wave energy sector remains in a nascent stage, constrained by several technical and economic barriers. A primary obstacle is the accurate forecasting of wave farm power output, which is crucial for maintaining grid stability, ensuring investment viability, and facilitating the broader commercial deployment of wave energy technologies.

To address this challenge, the present study investigates the convergence of green technological innovation and wave energy conversion, with a specific emphasis on forecasting the power output of wave energy farms. The goal is to facilitate the seamless integration of wave-generated electricity into the existing power infrastructure, thereby supporting a stable and resilient energy supply.
To this end, we introduce a novel predictive framework for wave energy output, leveraging the spatial characteristics of Wave Energy Converter (WEC) locations. The proposed approach employs a hybrid sequential learning architecture—a Self-Attention-enhanced Convolutional Bi-directional Long Short-Term Memory (Bi-LSTM) network, augmented with an efficient hyperparameter optimisation algorithm. This methodology offers a robust and scalable solution for modelling the complex temporal-spatial dynamics of wave power generation.
The model is trained and validated on four real-world datasets from operational wave farms along the southern coastline of Australia, specifically in Adelaide, Sydney, Perth, and Tasmania.

The performance of the proposed hybrid sequential predictive model is benchmarked against ten widely adopted machine learning (ML) algorithms. Experimental evaluations reveal that our model consistently outperforms the comparative methods, achieving high coefficients of determination ($R^2$) of 91.7\% (Adelaide), 88.0\% (Perth), 82.8\% (Tasmania), and 91.0\% (Sydney). These results underscore the superior predictive accuracy of the proposed framework across diverse marine environments. Moreover, the model demonstrates enhanced robustness and generalisation capability in forecasting wave farm power output, surpassing the performance of conventional ML and state-of-the-art deep learning approaches. Collectively, these findings affirm the effectiveness of the proposed Self-Attention Convolutional Bi-LSTM model in delivering reliable and accurate power predictions for wave energy systems.

\vspace{1em} 

\end{abstract}




\maketitle
\section{Introduction}
The need to urgently address greenhouse gas emissions and achieve carbon neutrality has
become a shared responsibility, emphasised by the United Nations Sustainable Development
Goal~\cite{zhou2023worldwide}. Concrete actions are paramount in the pursuit of carbon
neutrality, with green innovation emerging as a crucial approach closely aligned with this
goal~\cite{zheng2024digital}. As societies and global supply chains strive to meet net-zero
targets, those leading in green innovation and mastering innovative technologies will play a
pivotal role in bridging the gap~\cite{dohale2024critical}. Recognising the pivotal role of green
innovation in achieving carbon neutrality using modern Artificial Intelligence (AI) techniques is
significant in exploring its implications and potential.

Energy generation with minimal environmental impact has become increasingly
challenging~\cite{tao2024assessing} in recent years, prompting governments to support
renewable energy systems, especially compared to non-renewable resources in today's energy
market. Wave energy~\cite{yasir2025comprehensive}, a concentrated form of solar energy, holds particular promise among
renewable energy sources. Ocean wave energy is not just efficient; it is one of the most efficient
renewable sources, offering a high potential for large-scale clean energy production that
surpasses other renewables like solar and wind by orders of magnitude. Wave energy
converters facilitate the conversion of wave energy into electricity, a complex task in harnessing
ocean wave energy~\cite{golbaz2022layout}. Predicting the power output of wave energy converters in a wave farm is
crucial for optimising their performance and reducing computational costs. Various models,
including physical-based and statistical models, have been developed to address this challenge~\cite{khan2022towards}.

Numerous studies have been conducted on modelling, optimising, and predicting Wave energy and wave energy converters (WEC) performance in the last decade, with researchers constantly seeking ways to improve their efficiency and reliability. Statistical analysis has been utilised to forecast wave power, reducing computational costs. One area of interest is the application of artificial intelligence (AI) in evaluating the production potential of energy systems, providing efficient forecasting in this
domain~\cite{bento2021ocean,ni2018prediction}.
A hybrid model was proposed by Liu et al.~\cite{liu2020prediction}, a combination of a radial basis function neural network (RBFNN) and genetic algorithms (GAs) to predict and optimise a scaled oscillating wave surge converter. They numerically investigated these types of converters
using the smoothed particle hydrodynamics method and random changes in nine design parameters and capture factors to assess energy conversion efficiency with high accuracy at 83.33\%. While this approach shows promise, it is important to note that RBFNNs are computationally expensive, and selecting the optimal number of functions requires time and effort.
A 30\% increase was achieved in power capture efficiency using A feedforward back-propagation artificial neural network (FBNN) by Li et al.~\cite{li2021development} that addressed non-causality in real-time wave energy control by online forecasting of future wave force. The impact of prediction errors on power extraction was investigated, showing that phase
error had a greater influence than amplitude error. Additionally, a relationship between power capture efficiency and control constraint was identified.

Another hybrid predictive model was proposed~\cite{ni2019integrated} consisting of the long short-term memory (LSTM) algorithm and principal component analysis (PCA) for predicting electrical power generation from a WEC. The results demonstrate the superior performance of this integrated model compared to LSTM alone and other ML models, the support vector machine (SVM), regression tree (RT), Gaussian process regression (GPR) and ensembles of trees (ET). The experiments highlight the significant impact of high-frequency oscillating waves
and long-term features on the model's accuracy, showcasing the proposed model's
effectiveness in handling time sequence data and high-frequency signals.
A comparative study~\cite{wang2024electric} developed in order to predict electric power generated by a two-body hinge-barge WEC using four ML algorithms: Back-propagation Neural Network (BPNN), Long Short-Term Memory (LSTM), Support Vector Machine (SVM), and Radial Basis Neural Network (RBFF). The results demonstrate that BPNN achieved a coefficient of determination (R2) of up to 0.99\% on the test dataset.

In recent work~\cite{neshat2024meta}, the impact of the ensemble learning model was emphasised and proposed a Meta-learner gradient boosting method (MLGB). The Meta wave learner was trained and validated using wave farm datasets from Australia's southern coast, and its performance was compared with 15 other ML methods. The experimental results demonstrate the proposed model's competitiveness, with accuracy ranging from 84.4\% to 90.3\% across different locations and improved robustness in predicting wave farm power output.
One of the important parameters in wave power prediction is the significant wave height ($H_{sig}$), which has led to the application of AI models in numerous research studies. Time series models have shown superior performance compared to other ML models like linear regression, support vector machines (SVM), NNs, etc.
For instance, Ali et al.~\cite{ali2019significant} developed a hybrid ML model, ICEEMDAN-ELM, for forecasting significant wave height ($H_{sig}$) in the eastern coastal zones of Australia. The model combined the improved complete ensemble empirical mode decomposition method and the adaptive noise method with the extreme learning machine (ELM) algorithm, incorporating lagged historical Hs data as predictors. Comparisons with other models showed that this hybrid
model outperforms them (such as random forest (RF)), providing accurate Hs forecasts. This model can be valuable for decision-making when designing reliable ocean energy converters.

Neural networks possess the capacity to autonomously uncover and depict hierarchical characteristics through the arrangement of interconnected nodes across multiple layers. This notable attribute enables them to grasp intricate and nonlinear associations within input data, rendering them potent models for prediction-oriented tasks. A particular investigation~\cite{deshmukh2016neural} demonstrated the efficacy of employing wavelet neural networks in enhancing the numerical forecasting of significant wave height and peak wave period in a case study focused on a coastal region adjacent to Puducherry, situated on India's east coast. By comparing the suggested prediction method to a classic neural network model trained solely on measured data from Puducherry, it was found that the former approach outperformed the latter, exhibiting sustained prediction performance of superior quality.

A considerable accuracy was achieved~\cite{ali2020near} in the near-real-time significant wave height forecast model using a hybrid approach combining multiple linear regression (MLR) with covariance-weighted least squares (CWLS) estimation and optimised to forecast Hsig values 30 minutes ahead. The trained and validated data was from the eastern coastal zones of Australia.
The MLR-CWLS model is compared to other models, including MARS, M5 Model Tree, and MLR, and the results showed that the hybridised MLR-CWLS model provides reliable forecasts of Hsig compared to the other models (R2=0.97).
Pujan et al.~\cite{pokhrel2022transformer} proposed a novel approach for forecasting significant wave heights by combining machine learning models, specifically transformer neural networks, with numerical methods such as differencing and WaveWatch III. The study conducted a case study using data from 92 buoys. The experimental results demonstrated that the proposed model achieved excellent performance, with a root mean square error of 0.231 m for forecasting significant wave heights two days in advance. Comparing the proposed method with
LSTM, CNN, and Transformer models, the proposed approach outperformed the others,
showing superior accuracy in predicting significant wave heights.
In another study, a hybrid deep learning model called CLSTM-BiGRU~\cite{ahmed2024hybrid} was recently proposed to accurately forecast $H_{sig}$ for WEC sites in Queensland, Australia. The model combines Convolutional Neural Networks (CNNs), LSTM networks, and Bidirectional Gated Recurrent Units (BiGRUs) to predict $H_{sig}$ at different time horizons.
Historical wave properties and $H_{sig}$ are utilised to train the hybrid model, presenting improved accuracy compared to benchmark ML models. However, the study did not compare the complexity and training runtime of the proposed model with other ML models.

To tackle the challenges mentioned above related to predicting the power output of wave farms, particularly considering the intricate hydrodynamic interplays between WECs and the inherent variability of wave conditions, precision in power output predictions remains a challenging task. To address these complexities, we introduce a robust and efficient hybrid sequential deep learning model. The primary contributions of this research endeavour are outlined as  follows:
\begin{itemize}
\item A comprehensive analysis is conducted, contrasting ten established sequential and hybrid machine learning methods to formulate a specialised comparative AI-based framework tailored to precisely anticipate power output in wave farms across diverse real wave scenarios.
\item Also, we propose an adept and precise hybrid model by combining a developed convolutional deep model with a multi-layer bidirectional long short-term memory network and a self-attention mechanism, thus fortifying the accuracy of predictions.
\item In response to the challenge posed by hyper-parameters and architecture optimisation, the proposed hybrid model implements a swift heuristic algorithm, merging grid search techniques with the 1+1 Evolutionary Algorithm (1+1EA).
\item Furthermore, the performance of the hybrid predictive model is meticulously evaluated utilising an extensive dataset reflective of four real wave scenarios, with multiple metrics employed to assess its efficacy.
\end{itemize}

Through these concerted efforts, this study aims to advance the state-of-the-art predictive modelling for wave farm power output,  fostering a deeper understanding of how advanced machine learning methodologies can be harnessed to enhance predicting precision and operational efficiency in green energy conversion technologies.

This study begins with an initial exposition on the description and modelling of the wave energy converter (WEC) system, encompassing the fundamental equations of motion governing its behaviour, as well as a detailed explanation of the chosen deployment site and the performance measures employed for evaluating its effectiveness (detailed in Section~\ref{sec:wave_modelling}). Subsequently, a variety of machine learning (ML) and deep learning algorithms are introduced, along with the intricate technical aspects of the proposed Meta ensemble deep learning method, outlined in Section~\ref{sec:method}. The numerical results obtained through the application of this method are then meticulously presented and discussed in Section~\ref{sec:results}, facilitating a comprehensive comparison of the efficiency and effectiveness of the proposed approach. Finally, the manuscript concludes by summarizing the key findings and highlighting the advantages offered by the proposed method over existing approaches (Section~\ref{sec:conclusion}).

\section{System description and modelling}
\label{sec:wave_modelling}
\begin{figure}[h!]
\centering
 \includegraphics[clip,width=0.6\columnwidth]{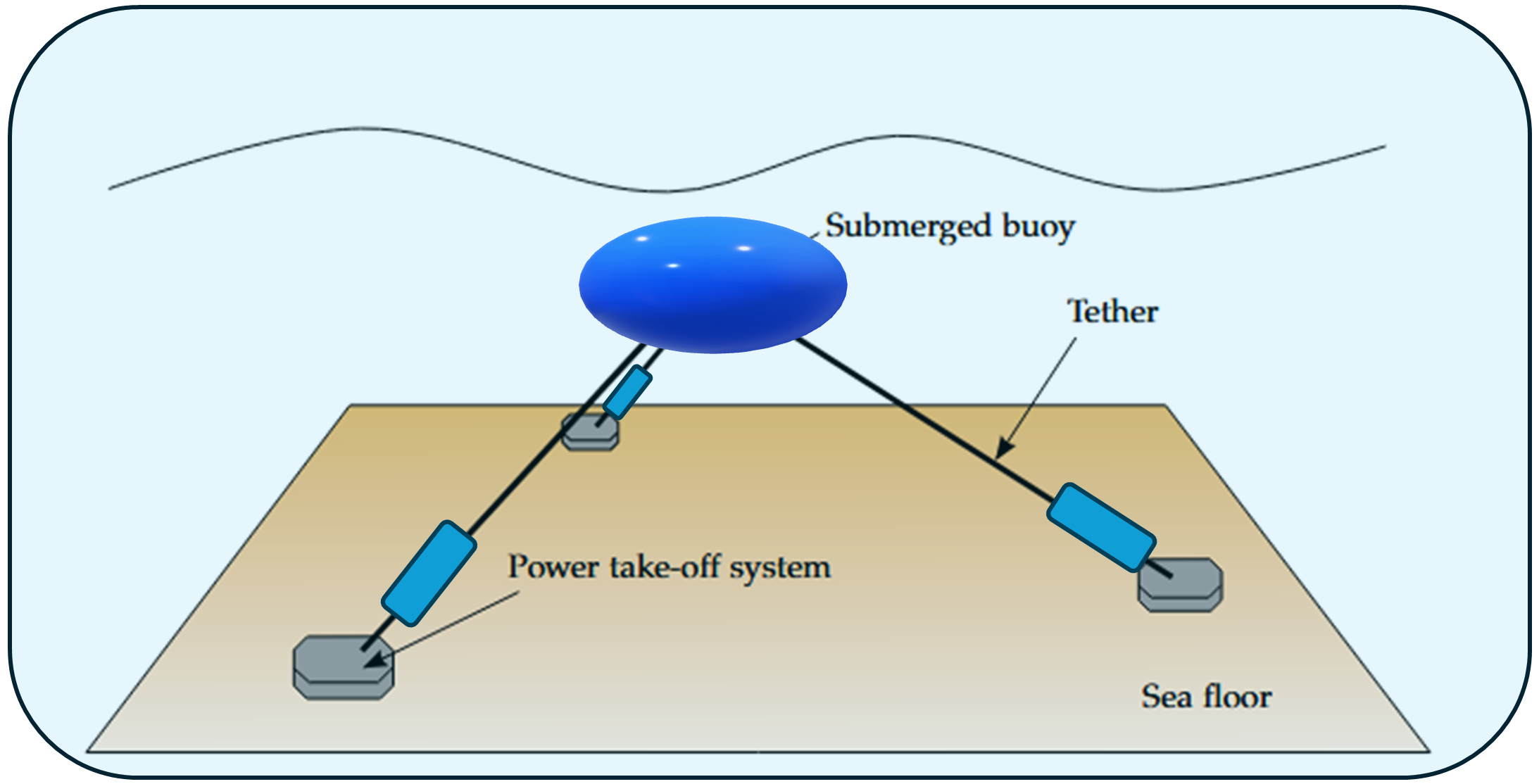}
 \caption{ schematic of a fully submerged three tethers wave energy converter.}
\label{fig:wave_farm}
\end{figure}

The utilisation of submerged spheres as wave power absorbers was initially proposed in the literature\cite{srokosz1979submerged} and has been implemented in the current study. The wave farm is composed of several Wave Energy Converters (WECs) positioned in proximity to one another. Each sphere has a radius of 5 meters, with the distance from the buoy's centre of mass to the still water level fixed at 8.5 meters. Three mooring lines are employed to anchor the sphere to the seabed. The mooring system's configuration is symmetrical to ensure that each WEC can independently absorb power regardless of the incoming wave direction. By inclining the mooring lines at 54 degrees relative to the vertical, the submerged sphere achieves optimal power absorption, as detailed in \cite{sergiienko2016optimal}. To maintain tension in the mooring lines, the buoys must be positively buoyant, necessitating that the buoy's mass be half that of the displaced volume. The wave farm comprises multiple closely situated WECs, as illustrated in Figure~\ref{fig:wave_farm}.
\subsection{Equations of motion}

The hydrodynamic loading on the Wave Energy Converters (WECs) is simulated using linear potential flow theory \cite{wu1995interaction}. The power take-off mechanism of each WEC is represented as a simplified linear spring-damper system with adjustable stiffness and damping control parameters. The dynamics of the WEC array are mathematically described in the frequency domain, considering wave directionality. The motion of each WEC is analyzed in the surge, sway, and heave modes, with the rotational movement of the spherical WECs being disregarded. The equations of motion are formulated as follows:

\begin{equation}
    \mathbf{x}(\omega,\beta) = \left(-\omega^2 \left(\mathbf{M} + \mathbf{A}(\omega)\right) + \rm{i}\omega \left(\mathbf{B}(\omega)+\mathbf{B}_{pto}\right) + \mathbf{K}_{pto} \right)^{-1}\mathbf{F}_{e}(\omega, \beta),
\end{equation}
In the given system, where $\omega$ represents the regular wave frequency, $\beta$ signifies the wave angle, $\mathbf{x}$ denotes the complex amplitude vector with dimensions of $[N\times3, 1]$, where $N$ stands for the count of WECs within a farm, $\mathbf{M}$ is the diagonal mass matrix, $\mathbf{A}$ and $\mathbf{B}$ are matrices encapsulating added mass and radiation damping coefficients respectively, accounting for hydrodynamic interactions among WECs in the farm. Moreover, $\mathbf{K}{pto}$ and $\mathbf{B}{pto}$ are matrices depicting power take-off stiffness and damping, governing the PTO mechanism, while $\mathbf{F}_e$ represents the wave excitation vector.

The assessment of the mean power absorbed by all WECs within a farm, contingent on the regular wave frequency and wave orientation, is determined as follows:

\begin{equation}
    \overline{P}(\omega, \beta) = \frac{\omega^2}{2}\mathbf{x}^{\rm{T}}(\omega, \beta) \mathbf{B}_{pto} \mathbf{x}(\omega, \beta),
\end{equation}

\subsection{Deployment sites and wave climates}

\begin{figure}[h!]
\centering
\subfloat[]{
 \includegraphics[clip,width=0.7\columnwidth]{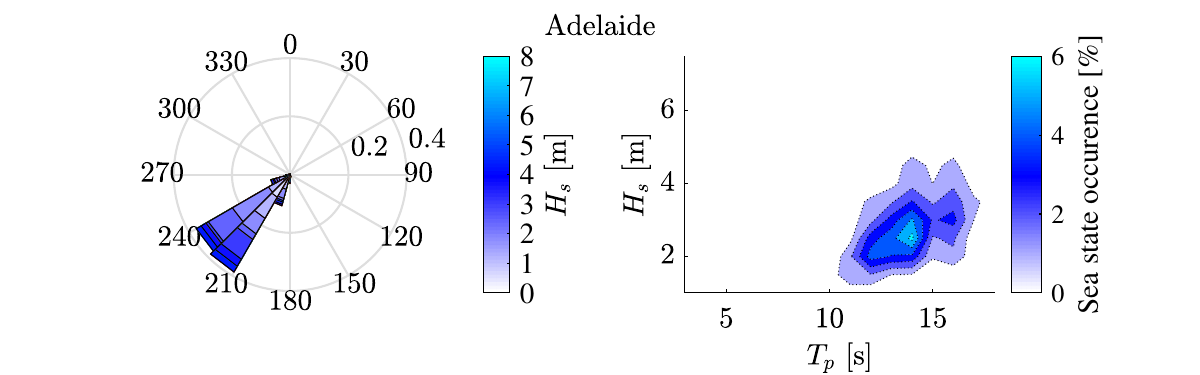}}\\
 \subfloat[]{
 \includegraphics[clip,width=0.7\columnwidth]{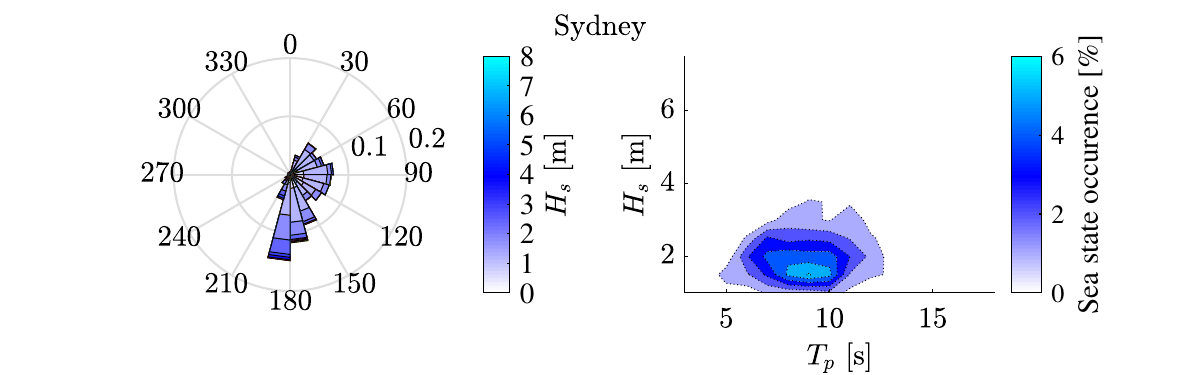}}
\caption{Wave scatter diagrams and the directional wave roses of four sites: (a) Adelaide, (b) Sydney.}
\label{fig:wave_climate}
\end{figure}

The efficiency of the wave farm is examined across four marine locations in Australia. The historical data for these specific sites is visualized in Figure ~\ref{fig:wave_climate}, sourced from the Australian Wave Energy Atlas.

The deployment sites in Adelaide and Sydney exhibit distinct wave climate characteristics, which are critical in evaluating the performance and generalisability of wave energy forecasting models. As illustrated in the wave rose and joint probability distribution plots, Adelaide's wave climate is predominantly influenced by southwesterly swells, with significant wave heights ($H_s$) mainly ranging between 1.5 and 4.5 meters, and peak wave periods ($T_p$) concentrated around 9 to 13 seconds. The sea state occurrence in Adelaide shows a relatively high density of energetic wave conditions, indicating its strong potential for wave energy extraction. In contrast, Sydney is characterized by a broader directional spread, primarily from the southeast, with lower wave energy levels. Here, $H_s$ typically remains below 3 meters, and $T_p$ is generally between 7 and 11 seconds, with a more moderate sea state occurrence distribution. These variations in wave dynamics between the two locations provide valuable insights into the spatial diversity of marine energy resources along the southern Australian coast and underscore the need for site-specific modeling approaches to accurately forecast power output.

\subsection{Performance measures}

Utilising frequency domain data, it becomes feasible to assess the potential power absorption capacity of a wave farm within an irregular wave environment, defined by parameters such as the significant wave height $H_s$, peak wave period $T_p$, and wave angle $\beta$:

\begin{equation}
    P_i(H_s, T_p, \beta) = 2\int _0^{\infty} S_i(\omega) \overline{P}(\omega, \beta) \rm{d}\omega
\end{equation}
where $S_i$ is the wave spectrum (Bretschneider in this study).

After computing the power output of the wave farm for each $i$-th sea state, the average annual power generation can be approximated by considering the probability of each wave condition's occurrence, denoted as $O_i(H_s, T_p, \beta)$:

\begin{equation}
    P_{AAP} = \sum_i^{N_s}P_i(H_s, T_p, \beta)\cdot O_i(H_s, T_p, \beta)
\end{equation}

\section{Methods}
\label{sec:method}
Recurrent neural networks~\cite{salehinejad2017recent} (RNNs) are an efficient and robust sort of neural network and algorithm with internal storage, making them one of the most intriguing algorithms. RNNs, like many other deep learning techniques, are relatively new. Despite being designed in the 1980s, it has only recently demonstrated its actual capabilities. With internal storage, the RNN can memorize significant aspects of the information it processes, allowing it to anticipate what will happen next with high accuracy. As a result, they are the finest algorithms for time series, voice, text, financial data, audio, video, and weather. Recurrent neural networks, as opposed to other algorithms, can provide a more in-depth comprehension of sequences and their surroundings.
The information in an RNN is cycled in a loop. When making a decision, it takes into account both the input information and what it has learned from prior inputs. The vanishing gradient problem happens when the gradient decreases through time as it rises. As a result, the layer that receives a modest gradient does not learn and provides the network with short-term memory. However, using LSTMs as a solution to the problem is another option~\cite{neshat2023short}.

LSTM~\cite{hochreiter1997long} is a sort of recurrent neural network (RNN) that is used to model long-distance sequences. It is a specialized RNN designed to address the gradient vanishing issue. A technique known as a gate allows LSTM to learn long-term reliance. These gates can learn which information in the sequence should be kept or discarded. The LSTM is comprised of three gates: input, forgetting, and output. In summary, the Forget gate chooses which data from the preceding phase should be saved. The input gate selects the necessary information to be added from the current stage. The next concealed state is determined by the output gate.
According to the research on the predictions analyzed, we opted to construct and evaluate nine distinct types of LSTM models and a CNN to estimate the power production of a wave farm. Each model has different unique properties. They are as follows: Vanilla LSTM, Stacked LSTM, Bi-directional LSTM, Stacked Bi-LSTM, Gated recurrent unit (GRU), Stacked GRU, CNN, CNN-LSTM, CNN-GRU, CNN-BiLSTM.

\subsection{Stacked LSTM}

Stacked LSTM networks are a type of recurrent neural network (RNN) that consists of multiple LSTM layers stacked on top of each other. Each layer captures different levels of abstraction and temporal dependencies in the input data, allowing the network to learn even more complex patterns in sequential data. The mathematical equations related to stacked LSTM networks can be summarized as follows. The forward LSTM layer processes the input sequence from left to right. The equations for the forward LSTM are similar to those of the standard LSTM but with the indices unchanged. The backward LSTM layer processes the input sequence from right to left. The equations for the backward LSTM are similar to those of the standard LSTM but with the indices reversed. Combining Forward and Backward LSTM Outputs: The final output of the BiLSTM network is the weighted sum of the forward and backward LSTM outputs.
\begin{equation}
o_t=\alpha o_{t, f}+(1-\alpha) o_{t, b}
\end{equation}

\noindent where $o_t$ is the final output, $\alpha$ is the weight given to the forward LSTM output, and $o_{t, f}$ and $o_{t, b}$ are the forward and backward LSTM outputs, respectively.
Stacked LSTM networks consist of multiple LSTM layers stacked on top of each other. The output of each LSTM layer is fed as input to the next LSTM layer. The equations for the stacked LSTM can be written as follows:
\begin{equation}
h_t^{(l)}=\operatorname{LSTM}_l\left(h_t^{(l-1)}\right)
\end{equation}

\noindent where $h_t^{(l)}$ is the hidden state of the $l_{t h} LSTM$ layer at time step $t$, and $LSTM_l$ is the $l_{t h} LSTM$ layer.
These equations describe the flow of information through a stacked LSTM network, which allows it to capture complex temporal dependencies in sequential data. Stacked LSTM networks have been demonstrated to be effective in various natural language processing tasks, including language modeling, machine translation, and speech recognition.

\begin{figure}[h!]
\centering
\includegraphics[width=0.7\linewidth]{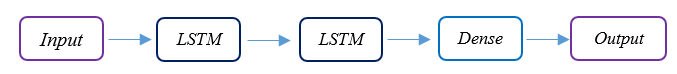}
\caption{The stacked long short-term memory (LSTM) architecture}
\label{fig:LSTM_st}
\end{figure}

\subsection{Bi-directional LSTM}

A Bidirectional Long Short-Term Memory~\cite{schuster1997bidirectional} (BiLSTM) network is a type of recurrent neural network (RNN) that learns bidirectional long-term dependencies between time steps of time-series or sequence data. BiLSTM networks are widely used in various applications, such as natural language processing, speech recognition, and time series prediction~\cite{joseph2023near}. The mathematical equations related to BiLSTM architecture can be summarized as follows:

Forward LSTM: The forward LSTM layer processes the input sequence from left to right. The equations for the forward LSTM are as follows:

Forget Gate:
\begin{equation}
f_{t, f}=\sigma_g\left(W_{f x_t}+U_f h_{t-1, f}+b_f\right)
\end{equation}

\noindent where $f_{t, f}$ is the forget gate, $W_f, U_f$, and $b_f$ are the corresponding weight matrices and biases, and $\sigma_g$ is the sigmoid activation function.

Input Gate:
\begin{equation}
i_{t, f}=\sigma_g\left(W_{i x_t}+U_i h_{t-1, f}+b_i\right)
\end{equation}

\noindent where $i_{t, f}$ is the input gate, $W_i, U_i$, and $b_i$ are the corresponding weight matrices and biases, and $\sigma_g$ is the sigmoid activation function.

Output Gate:
\begin{equation}
o_{t, f}=\sigma_g\left(W_{o x_t}+U_o h_{t-1, f}+b_o\right)
\end{equation}

\noindent where $o_{t, f}$ is the output gate, $W_o, U_o$, and $b_o$ are the corresponding weight matrices and biases, and $\sigma_g$ is the sigmoid activation function.

Cell State:
\begin{equation}
c_{t, f}=f_{t, f} \odot c_{t-1, f}+i_{t, f} \odot x_t
\end{equation}

\noindent where $c_{t, f}$ is the cell state, and $\odot$ denotes the element-wise multiplication.

Hidden State:
\begin{equation}
h_{t, f}=o_{t, f} \odot h_{t-1, f}+o_{t, f} \odot c_{t, f}
\end{equation}

\noindent where $h_{t, f}$ is the hidden state, $o_{t, f}$ is the output gate, and $h_{t-1, f}$ is the previous hidden state.

Backward LSTM: The backward LSTM layer processes the input sequence from right to left. The equations for the backward LSTM are similar to the forward LSTM, but with the indices reversed (e.g., $t$ becomes $t+1, h_{t-1, f}$ becomes $\left.h_{t+1, f}\right)$

Combining Forward and Backward LSTM Outputs: The final output of the BiLSTM network is the weighted sum of the forward and backward LSTM outputs.
\begin{equation}
o_t=\alpha o_{t, f}+(1-\alpha) o_{t, b}
\end{equation}

\noindent where $o_t$ is the final output, $\alpha$ is the weight given to the forward LSTM output, and $o_{t, f}$ and $o_{t, b}$ are the forward and backward LSTM outputs, respectively.
These equations describe the flow of information through the BiLSTM network, which allows it to capture bidirectional long-term dependencies in sequential data. The BiLSTM network's ability to learn from both the past and future components of an input sequence makes it a powerful tool for various applications in natural language processing and time series prediction.

\begin{figure}[h!]
\centering
\includegraphics[width=0.5\linewidth]{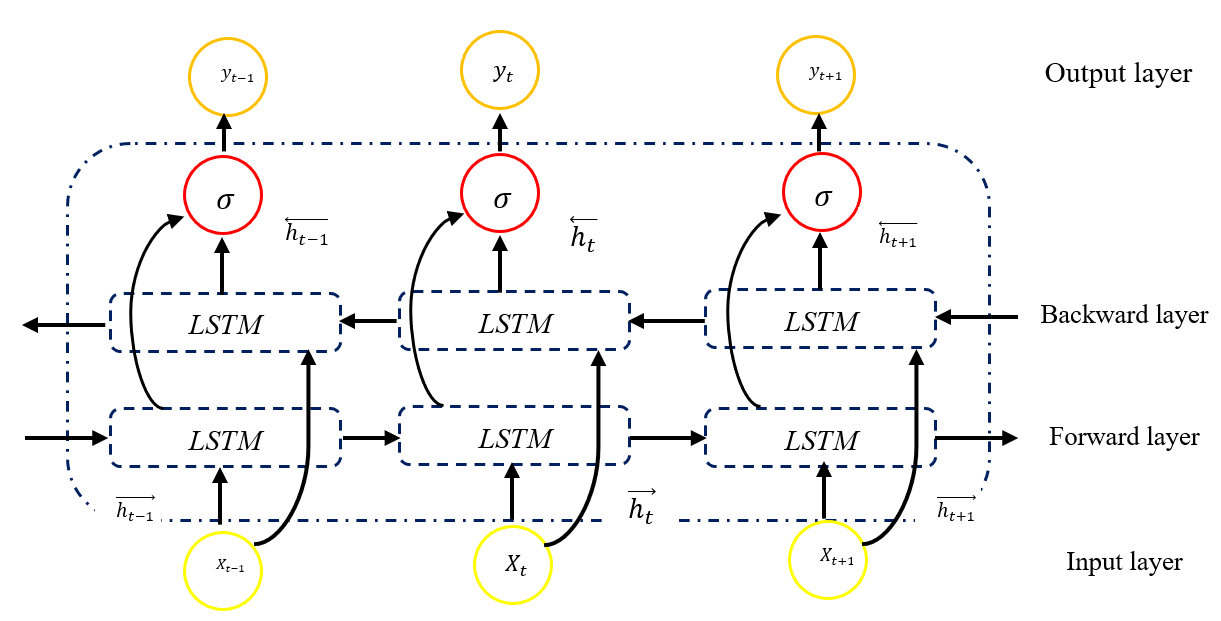}
\caption{The architecture of basic Bi-directional LSTM (BiLSTM).}
\label{fig:BiLSTM_st}
\end{figure}

\subsection{Gated recurrent unit (GRU)}

A Gated Recurrent Unit (GRU)~\cite{cho2014learning} network is a type of Recurrent Neural Network (RNN) that addresses the issue of long-term dependencies in sequential data by using gating mechanisms to selectively update the hidden state of the network at each time step. The mathematical equations related to GRU networks can be summarized as follows:

Update Gate: The update gate decides which information should be passed to the output and which information should be discarded. It is defined as follows:
\begin{equation}
z_t=\sigma_g\left(W_{z x_t}+U_z h_{t-1}+b_z\right)
\end{equation}

\noindent where $z_t$ is the update gate, $W_z, U_z$, and $b_z$ are the corresponding weight matrices and biases, and $\sigma_g$ is the sigmoid activation function.

Reset Gate: The reset gate, often referred to as the barrier gate, is responsible for resetting the hidden state to zero at specific points in the sequence. It is defined as follows:
\begin{equation}
r_t=\sigma_g\left(W_{r x_t}+U_r h_{t-1}+b_r\right)
\end{equation}

\noindent where $r_t$ is the reset gate, $W_r, U_r$, and $b_r$ are the corresponding weight matrices and biases, and $\sigma_g$ is the sigmoid activation function [2].
Candidate Hidden State: The candidate hidden state, $h_t^{(.)}$, is the intermediate memory state before the update. It is defined as follows:
\begin{equation}
h_t^{(.)}=\tanh \left(W_{h x_t}+U_h h_{t-1}+b_h\right)
\end{equation}

\noindent where $h_t^{(.)}$is the candidate hidden state, $W_h, U_h$, and $b_h$ are the corresponding weight matrices and biases, and tanh is the hyperbolic tangent activation function.
The hidden state, $h_t$, is updated using the update gate and the candidate hidden state as follows:
\begin{equation}
h_t=\left(1-z_t\right) \odot h_{t-1}+z_t \odot h_t^{(.)}
\end{equation}

\noindent where $h_t$ is the hidden state, and $\odot$ denotes the element-wise multiplication.
These equations describe the flow of information through a GRU network, which allows it to model sequential data effectively. GRU networks have been shown to be effective in various natural language processing tasks, such as language modelling, machine translation, and speech recognition.

\begin{figure}[h!]
\centering
\includegraphics[width=0.6\linewidth]{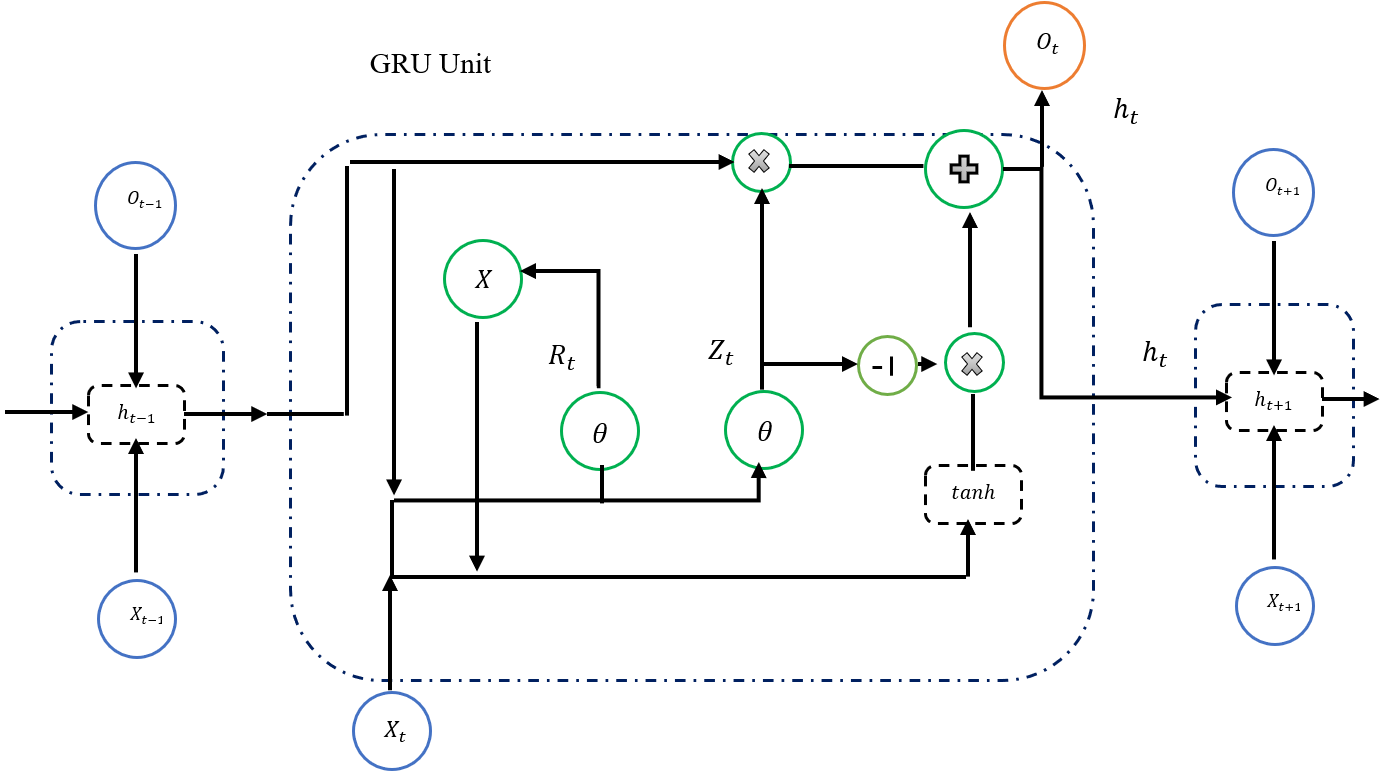}
\caption{The architecture of basic Gated Recurrent Unit (GRU).}
\label{fig:GRU_st}
\end{figure}

\subsection{Self-attention model}
The attention mechanism~\cite{shaw2018self}, a technique used in neural networks, assigns weights to different input components based on their importance and combines them to create a final representation. This allows the neural network to prioritize important input parts, leading to improved model performance.
The SE Block (Squeeze and Excitation Block) is a structural unit that efficiently focuses on the channel dimension and incorporates a channel attention mechanism into the model. By assigning weights to individual feature channels, the SE Block efficiently enhances or suppresses specific channels for different tasks, enabling the extraction of valuable features. The  SE attention mechanism involves several steps, including compressing two-dimensional features into real numbers using global average pooling, generating weight values for each feature channel through fully connected layers, and applying normalised weights to the channel features. These steps collectively enhance the model's predictive accuracy by adjusting the importance of different channel features.
\subsection{Proposed hybrid model: CNN-BiLSTM-SA with adaptive hyper-parameter optimiser}

The primary limitation of fully connected LSTM and BiLSTM when dealing with spatiotemporal data lies in its reliance on full connections for input-to-state and state-to-state transitions~\cite{duan2022combined}, lacking the incorporation of spatial information. Moreover, it treats all input features equally, disregarding the sequence's spatial relationships between distinct components. Furthermore, In tabular data such as wave energy converter datasets, the spatial relationships are not as clear, making it harder for CNNs to extract meaningful features. Also, wave energy converter datasets often contain diverse features with complex interactions and dependencies. CNNs may not naturally capture these inter-feature relationships as effectively as other architectures. 

To address the above shortages of LSTM, BiLSTM and CNNS, various solutions were suggested~\cite{shi2015convolutional}; however, each one suffered from generalisation and robust ability to solve different problems with various characteristics. 
In this way, we proposed a hybrid model that stands out by featuring inputs ($X_1,...,X_t$), cell outputs ($C_1,...,C_t$), hidden states ($H_1,...,H_t$), and gates ($i_t,o_t,f_t$) in the hybrid CNN-BiLSTM as 3D tensors, with the last two dimensions representing spatial aspects. Through the CNN-BiLSTM, the prediction of a cell's future state within the grid hinges on its neighbouring cells' inputs and previous states, forming a cohesive spatially-aware framework.
The integration of CNN and BiLSTM models combines the advantages of both architectures to tackle the spatial and temporal complexities inherent in WEC layout datasets. CNNs are adept at extracting spatial features by utilising convolutional layers, which are proficient in identifying and learning patterns within spatial data, making them well-suited for processing the spatial coordinates of WECs.
On the other hand, BiLSTM networks are specifically designed to capture long-term dependencies in sequential data by processing information from both past and future contexts within a sequence, which is essential for understanding the temporal dynamics of wave energy data. By merging CNN and BiLSTM, the hybrid model is able to simultaneously capture spatial dependencies through CNN layers and temporal dependencies through BiLSTM layers, thereby ensuring a more comprehensive understanding and precise prediction of wave farm power output.

For each sequence, vector $D$, a convolutional operation is conducted to extract high-level features and a hyperbolic tangent linear activation function is employed. Ultimately, the output generated by the convolutional neural network is represented by a new vector $X = x_{1}\bigoplus x_{2}\bigoplus x_{3}\bigoplus ... \bigoplus x_{t}\bigoplus ...\bigoplus x_{k}$, which can be expressed as:
\begin{equation}
\boldsymbol{X}= f\left(\operatorname{conv}\left(\boldsymbol{W}_{c}, \boldsymbol{D}\right)\right)
\end{equation}

\noindent where $\boldsymbol{W}_{c}$ denotes the weight associated with convolution, and $f(0)$ represents the hyperbolic linear unit:

\begin{equation}
\boldsymbol{f(x)}= \begin{cases}x, & x>0 \\ \frac{\alpha x}{1-x}, & x \leq 0\end{cases}
\end{equation}
$\alpha$ is a hyper-parameter.

Following the execution of the CNN operation, the Bidirectional Long Short-Term Memory (BiLSTM)~\cite{schuster1997bidirectional}  with a hidden neuron size of $l$ is utilized to effectively capture the high-level contextual information, which is synthesized from multiple LSTM layers, specifically the forward LSTM and the backward LSTM. 
Subsequently, the output of the BiLSTM is represented as a $k\times nl$ dimensional vector, with the intricate details articulated in Equations (25)–(32), 
wherein $x_{t}$ signifies the input to the $t_{th}$ LSTM cell, $Oc_{lstm_{t-1}}$ represents the output from the preceding LSTM cell; $\sigma $ indicates the logistic 
regression function known as Sigmoid; $\ast$ denotes element-wise multiplication; and $W_{f}$, $W_{i}$, $W_{c}$, $W_{o}$, along with $b_{f}$, $b_{i}$, $b_{c}$, $b_{o}$, are designated as the weights and biases corresponding to the $t_{th}$ LSTM cell.

\begin{equation}
\boldsymbol{FG}_t=\sigma\left(\boldsymbol{W}_f\left[\boldsymbol{OC}_{lstm-1}, \boldsymbol{x}_t\right]+\boldsymbol{b}_f\right),
\end{equation}
where $\boldsymbol{FG}_t$ denotes the forget gate within an LSTM cell, which possesses the capability to ascertain the specific information that necessitates elimination via $\boldsymbol{x}_t$ and $\boldsymbol{OC}_{lstm-1}$.

\begin{equation}
\boldsymbol{ig}_t=\sigma\left(\boldsymbol{W}_i\left[\boldsymbol{OC}_{lstm-1}, \boldsymbol{x}_t\right]+\boldsymbol{b}_i\right)
\end{equation}

\begin{equation}
\tilde{\boldsymbol{c}}_t=\tanh \left(\boldsymbol{W}_c\left[\boldsymbol{OC}_{lstm-1}, \boldsymbol{x}_t\right]+\boldsymbol{b}_c\right),
\end{equation}

$\boldsymbol{ig_{t}}$ represents the input gate of the LSTM cell, possessing the authority to determine which information ought to be modified through $\boldsymbol{x}_t$ and $\boldsymbol{OC}_{lstm-1}$; subsequently, the candidate cell information $\boldsymbol{\hat{c}_{t}}$ can be derived as articulated in Equation (27).

\begin{equation}
\boldsymbol{ci_{t}} = \boldsymbol{FG}_{t} * \boldsymbol {ci}_{t-1} + \boldsymbol {ig}_t * \boldsymbol{\hat{c}_{t}}
\end{equation}

$\boldsymbol {{ci}_{t-1}}$ signifies the previous cell information, which subsequently influences the decision regarding the information to be discarded through the forget gate, while $\boldsymbol{\hat{c}_{t}}$ dictates the information to be refreshed via the input gate, culminating in the derivation of the new cell information $\boldsymbol{ci_{t}}$.

\begin{equation}
\boldsymbol{OG}_t=\sigma\left(\boldsymbol{W}_o\left[\boldsymbol{OC}_{lstm-1}, \boldsymbol{x}_t\right]+\boldsymbol{b}_o\right),
\end{equation}

\begin{equation}
\boldsymbol{OC_{lstm}}=\boldsymbol{OG}_t * \tanh \left(\boldsymbol{ci}_t\right)
\end{equation}

$\boldsymbol{OG}_t$ constitutes the output gate of the LSTM cell, which is capable of evaluating the conditions pertinent to the output based on $\boldsymbol{x}_t$ and $\boldsymbol{OC_{lstm}}$, thereby allowing for the generation of the LSTM cell output $\boldsymbol{OC}_lstm$ through the output gate in conjunction with the cell information.
Ultimately, the resultant output of BiLSTM $\boldsymbol{O}$ can be denoted as:

\begin{equation}
\boldsymbol{O}=\boldsymbol{o}_1 \oplus \boldsymbol{o}_2 \oplus \cdots \oplus \boldsymbol{o}_t \oplus \cdots \oplus \boldsymbol{o}_k
\end{equation}
\begin{equation}
\boldsymbol{o}_t=\left[\boldsymbol{OC}_{lstm}^f, \boldsymbol{OC}_{lstm}^b\right]
\end{equation}

\noindent where $\boldsymbol{OC}_{lstm}^f$ represents the output generated by the $\boldsymbol{t_th}$ cell within the forward Long Short-Term Memory (LSTM) architecture, and $\boldsymbol{OC}_{lstm}^b$ signifies the output produced by the $\boldsymbol{t_th}$ cell within the backward LSTM framework.

These equations describe the flow of information through the BiLSTM network, which allows it to capture bidirectional long-term dependencies in sequential data. 

Furthermore, the self-attention mechanism ~\cite{shaw2018self} is employed to allocate varying weights to the output features generated by the BiLSTM, with the resultant output being regarded as the single-channel coding features. This allows the BiLSTM to prioritize important input parts, leading to improved model performance. The SE Block (Squeeze and Excitation Block) is a structural unit that efficiently focuses on the channel dimension and incorporates a channel attention mechanism into the model. By assigning weights to individual feature channels, the SE Block efficiently enhances or suppresses specific channels for different tasks, enabling the extraction of valuable features. The specifics are enumerated as follows:


\begin{equation}
\boldsymbol{F}=\tanh \left(\boldsymbol{S}_k \boldsymbol{O}^T\right)
\end{equation}
\begin{equation}
\begin{aligned}
& \boldsymbol{A}=\operatorname{softmax}\left(\boldsymbol{S}_a \boldsymbol{F}\right), \\
& \boldsymbol{O_{AL}}=\boldsymbol{A O}
\end{aligned}
\end{equation}

\noindent where $\boldsymbol{O}^T$ represents the transposed matrix of $O$, $\boldsymbol{O_{AL}}$ indicates the resultant output of the attention layer, and $\boldsymbol{S}_k$, $\boldsymbol{S}_a$ are parameters that are subject to training. The preceding exposition has elaborated on the conventional architecture of a single channel. 

The incorporation of self-attention with CNN and BiLSTM layers is anticipated to yield several significant enhancements, including heightened predictive accuracy, improved generalisation across diverse sea sites, and more efficient training through adaptive hyper-parameter tuning.
The advantages of the combination of these methods can be listed as follows:
\begin{itemize}
    \item The incorporation of CNN and BiLSTM enables the model to comprehend both spatial and temporal relationships, resulting in notably improved prediction accuracy in comparison to individual models.
    \item Adaptability and Versatility: The model's structure can be readily adjusted to different wave farm setups and datasets, making it a versatile tool for predicting wave energy. 
    
\end{itemize}

\begin{figure}[h!]
\centering
 \includegraphics[clip,width=1.05\columnwidth]{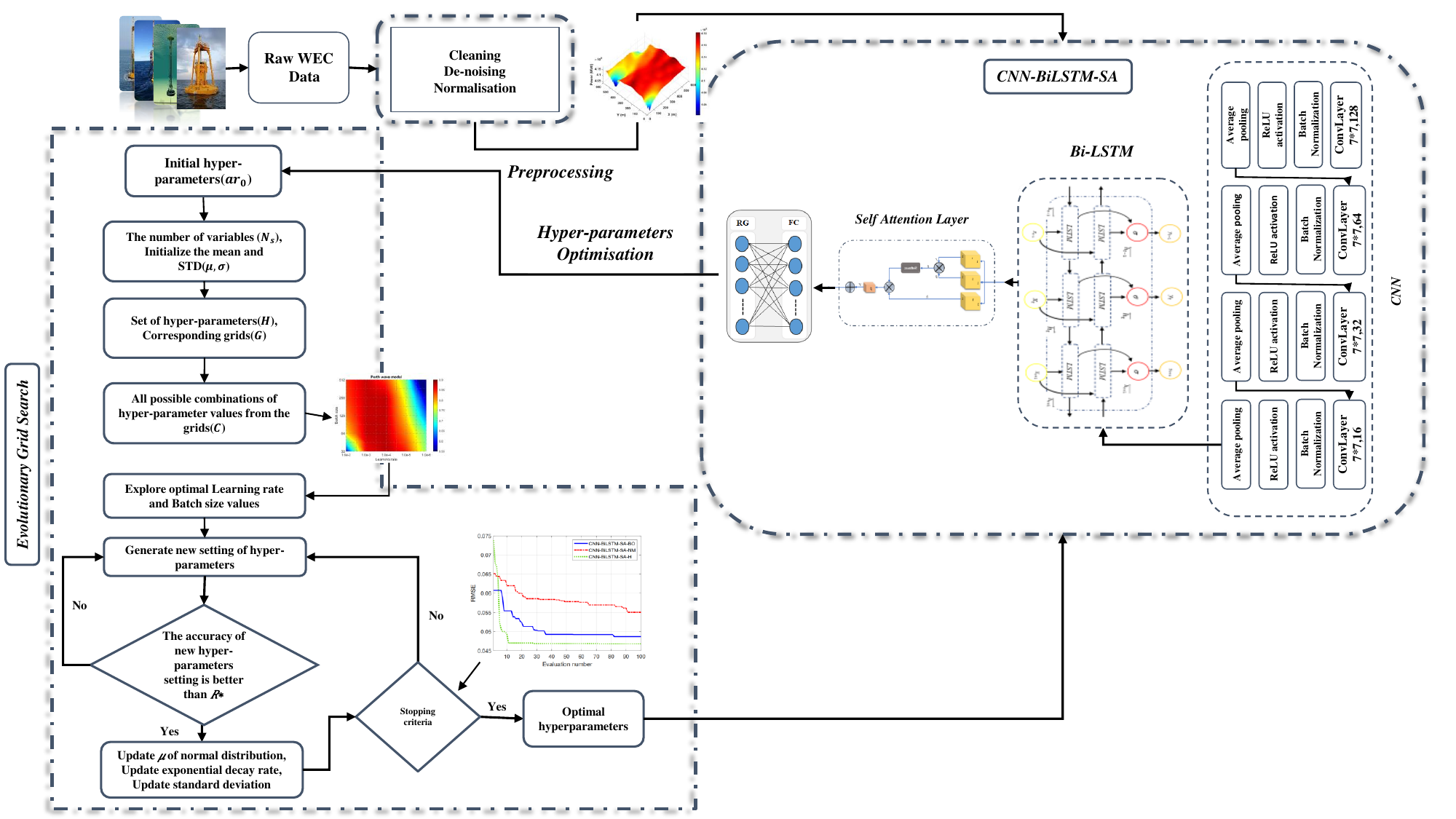}
 \caption{ The proposed hybrid deep model: CNN-BiLSTM-SA with hyper-parameters tuning method (EGS)   }
\label{fig:proposed_model}
\end{figure}

The proposed hybrid CNN and BiLSTM model, enriched with a self-attention mechanism and adaptive hyper-parameter optimization, signifies notable progress in forecasting the power output of wave farms. By integrating spatial and temporal feature extraction capabilities with targeted attention on significant data points, this model addresses the challenges associated with wave energy power prediction, thus facilitating more efficient integration of wave energy into the power grid. Figure~\ref{fig:proposed_model} demonstrates a schematic of various components of the proposed hybrid model. 

\subsubsection{ Tuning the Hyper-parameters of deep neural networks  }

Hyper-parameters refer to a collection of parameters utilized in the training and testing stages of the learning process. When training a deep learning model, it is essential for the model to learn various instances and weights that correspond to different feature combinations as patterns~\cite{hanifi2024advanced}. Common hyper-parameters consist of the learning rate, number of iterations, hidden layers, batch size, activation functions, momentum, and regularization. In the context of convolutional neural networks used for image classification, factors such as the field size of convolutional and pooling layers and the stride parameter for handling step size are taken into account. These parameters can be represented as integers or continuous or categorical variables within specified lower and upper bounds. Intermediate layers may introduce additional hyper-parameters, leading to potential variation in the number of neurons across layers.

During the training process, it is essential to select the optimal hyper-parameters to effectively train the model, as this can significantly impact its performance. Inadequate learning rates may inadvertently omit crucial patterns. The combination of parameters should be carefully chosen to minimize the loss function or maximize the model's performance and accuracy. Tuning hyper-parameters can be viewed as an optimization problem akin to fuzzy optimization~\cite{angelov1994generalized,baruah2013dec} applied in raw data. The hyperparameters of a deep learning model remain constant throughout training, enhancing model accuracy while also considering memory and time constraints to reduce the loss function value.
The selection of hyper-parameters is specific to each model and is determined by the problem being addressed. There is no universally optimal set of hyper-parameters applicable to all models~\cite{michael2024cohesive}. This study has focused on the learning rate, momentum, regularization, and network depth. The learning rate aids in recognizing general patterns in images, while momentum assists in thoroughly exploring the entire search space without overlooking important points. Regularization supports the model in achieving better generalization, thereby preventing overfitting and enhancing prediction accuracy.

\subsubsection{Bayesian optimization}
Deep learning optimization involves solving black-box optimization problems where the objective function, denoted as $f(y)$, is not explicitly known. It is crucial to minimize the number of samples used across the layers. In cases where human expertise cannot significantly contribute to improving accuracy, Bayesian optimization proves to be highly beneficial. This approach incorporates prior information about the function $f$ and continually updates posterior information, leading to reduced loss and enhanced model accuracy and has successful application in optimising renewable energy systems~\cite{baheri2018iterative}.

Bayesian optimization efficiently uncovers the global optima of the black box function within neural networks. It effectively handles noisy data and non-continuous spaces to achieve global minima while making optimal use of available resources.

The optimization process relies on Bayes' Theorem as described by~\cite{kramer2011derivative} in the equation that pertains to a model $Z$ and an observation $Y$.
\begin{equation}
P(Z \mid Y)=(P(Y \mid Z) P(Z)) / P(Z)
\end{equation}

Bayesian optimization involves finding the minimum of a function $f(y)$ within a bounded set $Y$, using the posterior probability $P(Z \mid Y)$, likelihood $P(Y \mid Z)$, prior probability $P(Z)$, and marginal probability $P(Z)$.

Although Bayesian optimization presents a realm of efficient exploration, reduced function evaluations, and adeptness in navigating through noisy functions and diverse objectives, it can pose a challenge with its computational intricacy, susceptibility to hyper-parameters, restrictions in scaling up in high-dimensional realms, and susceptibility to biases in initialization. 
In this paper, the Root Mean Square Error (RMSE) was used as an objective function for the Bayesian optimization algorithm to optimize the proposed model’s hyper-parameters. 


\subsubsection{Nelder-mead optimisation}
The pioneering Nelder-Mead approach is widely acclaimed as the foremost direct search technique for optimising unconstrained objective functions. This strategy involves assessing function values at the $n+1$ vertices $x_i$ of a simplex, where $n$ indicates the problem's dimensionality. The simplex, initially positioned at $x_0$ randomly, adheres to a particular guideline for determining its magnitude, as stipulated by~\cite{barthelemy1993approximation}. By continuously comparing function values and adjusting the simplex, the Nelder-Mead method strives to move closer to the problem's optimal solution~\cite{luersen2004constrained}.
\begin{equation}
x_i=x_0+p e_i+\sum_{\substack{k=1 \\ k \neq i}}^n q e_k, \quad i=1, n: => 
  \begin{cases} 
   p=\frac{a}{n \sqrt{2}}(\sqrt{n+1}+n-1) &   \\
   q=\frac{a}{n \sqrt{2}}(\sqrt{n+1}-1)       &  
  \end{cases}
\end{equation}

where $e_i$ are the unit base vectors. 
Throughout the implementation of the NM technique, the simplex's vertices are consistently altered through \textit{reflection}, \textit{expansion}, and \textit{contraction} manoeuvres. These manoeuvres are designed to pinpoint a better configuration that steers the algorithm towards the most favourable solution. The sequence persists until the values of the simplex vertices reach a level of similarity that signals convergence, which is assessed through an inequality standard as follows. 
\begin{equation}
\sqrt{\sum_{i=1}^{n+1}\left(f_i-\bar{f}\right)^2 / n}<\varepsilon, \bar{f}=\frac{1}{n+1} \sum_{i=1}^{n+1} f_i
\end{equation}

\noindent where $\varepsilon$ dances as a tiny optimistic scalar, the combined impact of the actions performed on the simplex can be likened to elongating the form in the paths of descent and swiftly navigating around nearby peaks. 
In a nutshell, the NM optimisation technique provides a blend of simplicity, resilience, and adaptability for enhancing various functions. Nevertheless, it could face challenges such as sluggish convergence, absence of assured global convergence, vulnerability to initial conditions, and restricted scalability in complex spaces with many dimensions. 

\subsubsection{Evolutionary Grid Search}

Widely adopted strategies like local and global optimisation methods involve tuning hyper-parameters and architectures to enhance the performance of predictive ML and deep learning models. This task proves challenging and time-consuming due to the vast number of possible variable combinations. A solution set, denoted as $X$, is defined alongside a real-valued function, $f$, mapping $X$ to real numbers. The objective is to explore an optimal solution, $x^*$, where $x^* \in X$ with $f(x^*) \le f(x), \forall x \in X$. As these combinatorial problems grow exponentially with expanded variables, a rapid and efficient optimisation method becomes necessary to converge towards a suitable solution while minimising the number of evaluations required.
However, local search methods encounter the drawback of being susceptible to local optima, restricting their ability to explore substantial portions of the search space and hindering their capability to discover the global optimum in multimodal problems. Conversely, global search methods, such as evolutionary algorithms, exhibit slow convergence due to the necessity for numerous evaluations to attain a satisfactory solution.

In this paper, a fast and adaptive Evolutionary Grid Search (EGS) algorithm is proposed to develop the performance of the proposed hybrid forecasting model by tuning the hyperparameters.  EGS is a combination of a grid search with an iterative local search (1+1EA) with an adaptive exploration step size in which a local optimum is in the neighbourhood of the assembled solution.  EGS has three steps, including $i)$ grid search, $ii)$ exploration, and $iii)$ exploitation phase.  In the first step, in order to speed up the optimisation process, we used a grid search algorithm and specified a set of ranges for both learning rate and batch size hyper-parameters, which play a significant role in training deep learning models.  After finding the optimal configuration of the learning rate and batch size, they are kept fixed.  Then, we selected another pair of hyper-parameters, such as the number of neurons and filters, and ran an evolutionary local search.  This procedure is followed to ensure all parameters are tuned.  The technical details of the EGS Algorithm~\ref{alg:EGS} can be seen in the following.   

\begin{algorithm}[h!]
\small
\caption{$\mathit{Evolutionary\, Grid\, Search (EGS)}$}\label{alg:EGS}
\begin{algorithmic}[1]
\Procedure{EGS ($solution_0={h_1,h_2,...,h_n}$ )}{}\\
 \textbf{Initialisation}
 \State $\mathit{ar_0}=\mathit{Eval(solution_0)}$ \Comment{Evaluate model by initial hyper-parameters}
\State $N_s=len(solution_0)$ \Comment{The number of variables}
\State $\mu=array$,$~\sigma={\sigma_1,\sigma_2,...,\sigma_{N_s}}$, $\lambda=-0.04$  \Comment Initialise the mean and STD

\textbf{ Grid Search }
\State $H = \{h_1, h_2, ..., h_{N_s}\}$ \Comment{Set of hyper-parameters}
\State $G = \{G_1, G_2, ..., G_{N_g}\}$ \Comment{Corresponding grids}
\State $C = (c_1, c_2, ..., c_{N_c}) \in G_1 \times G_2 \times ... \times G_{N_g}$ \Comment{All possible combinations of hyper-parameter values from the grids}
\For{ $i$ in $[1,..,N_{lr}]$ } \Comment{Explore optimal Learning rate and Batch size values}
\For{ $j$ in $[1,..,N_{bs}]$ }
\State $R(c_{i,j})= Train(Model(c_{i,j}))$
 \EndFor
  \EndFor 
  \State $<c^*,R^* >= argmax(R(c_{i,j})), where (c_{i,j}) \in C$ \Comment{Best-performed configuration of the Grid search}\\
\textbf{Adaptive 1+1EA}
\State $\mu = c^*$
\While {\textit{t != stopping criteria}}
 \For{ $i$ in $[1,..,N_s]$ }
 \If{$rand \le \frac{2}{N_s}$}
 $solution_i$ = $\mathcal{N}(\mu_i,\,\sigma_i^{2})$ \Comment{Generate new setting of hyper-parameters}
 \EndIf
  \EndFor 
 \State $\langle\mathit{R_t}\rangle$={\em{Train}}$(Model(\mathit{solution}))$
 
\If{$\mathit{R_t}> R^* $} \Comment{The accuracy of new hyper-parameters setting is better than $R^*$}
\State $\mu=solution$ \Comment {Update $\mu$ of normal distribution }
\State $\lambda=\lambda - r_d$ \Comment{Update exponential decay rate}
\State $\sigma=\sigma_0 \times e^{-\lambda t}$ \Comment{Update standard deviation}
\State $R^*=R_{t}$
\EndIf
   \EndWhile
\State \textbf{return} $solution,\mathit{R^*}$ \Comment{Optimal configuration}

\EndProcedure
\end{algorithmic}
\end{algorithm}
\section{Experimental results}
\label{sec:results}
The results obtained from the previously mentioned methods on our datasets are discussed in this section.

\subsection{Datasets and analysis}

In order to execute statistical modelling, a considerable amount of data needs to be available for analysis~\cite{neshat2019hybrid}, and to collect and publish the current model wave energy dataset. This dataset was obtained from UCI's machine learning library. The data set offers information on the positioning of WEC along Australia's southern coast in four natural wave cycles (Sydney, Adelaide, Perth, and Tasmania). The converter model used is a completely submerged three-tether converter known as CETO. In a space-constrained context, the 16 WEC spots are positioned and optimized. The first 32 variables of the data set are the positions (X coordinates and Y coordinates) of the 16 WEC on a continuous scale ranging from 0 to 566 (m) of the wave farm.
The following 16 attributes represent the absorbed power of each Wave Energy Converter (WEC) in the wave farm. The final attribute is the total power output from the wave farm, which is calculated as the sum of all 16 absorbed powers. The placement of WECs in relation to their generated power is used to predict the wave farm's power output. Each dataset consists of 72000 different combinations of WEC placements and their associated absorption powers. To assess the predictive model's accuracy, we split the dataset into 70\% test data and 30\% validation data. Moreover, to construct the predictive model, the data is normalized and scaled.

\subsection{Wave farm landscape analysis}

In the landscape analysis that was conducted, the focus was on exploring the relationship between the position adjustments of individual WECs and their impact on the total power output in the wave farms located in Adelaide and Sydney sea sites. Two specific layouts were selected for evaluation: one with four WECs and another with eight WECs. The objective was to understand how modifying the position of a single WEC while keeping the remaining WECs fixed would influence the overall power production.
To carry out the experiments, a systematic approach was employed. The positions of the $N-1$ buoys ($B1, B2,..., B_{N-1}$) were kept constant, representing the fixed layout, while the position of the last buoy ($B_N$) was iteratively adjusted. The separation distance between $B_N$ and the other buoys within the wave farm was varied across a range of distances, starting from a minimum of 50 m (refer to safe distance) and extending to all feasible locations. This strategy allowed for a comprehensive analysis of the impact of relative positions on power output.

\begin{figure}[h!]
\centering
\subfloat[]{
 \includegraphics[clip,width=0.25\columnwidth]{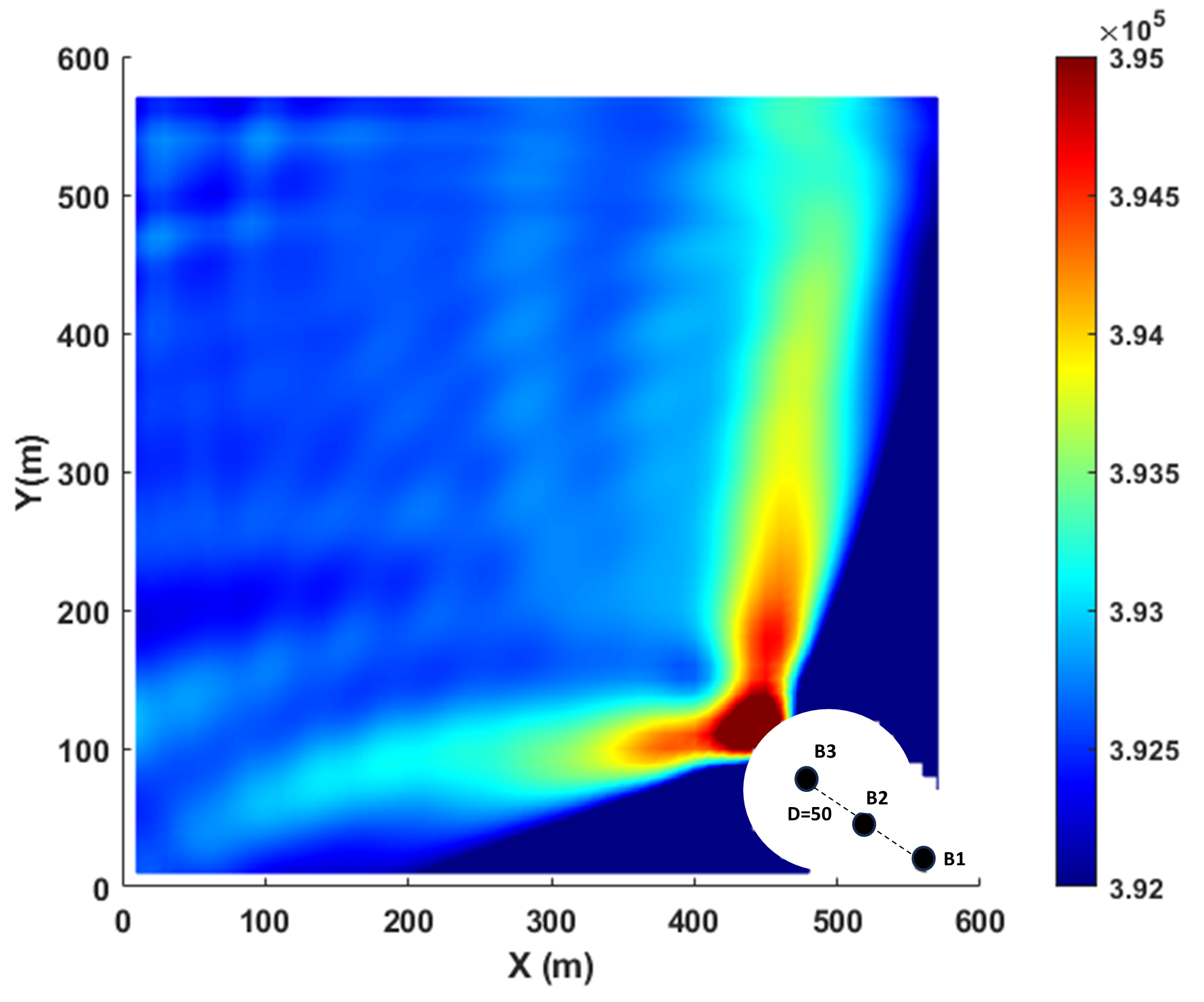}}
 \subfloat[]{
 \includegraphics[clip,width=0.25\columnwidth]{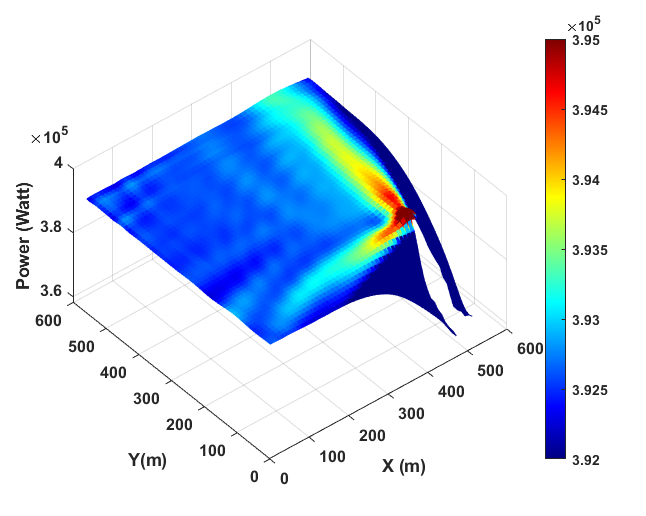}}
 \subfloat[]{
 \includegraphics[clip,width=0.25\columnwidth]{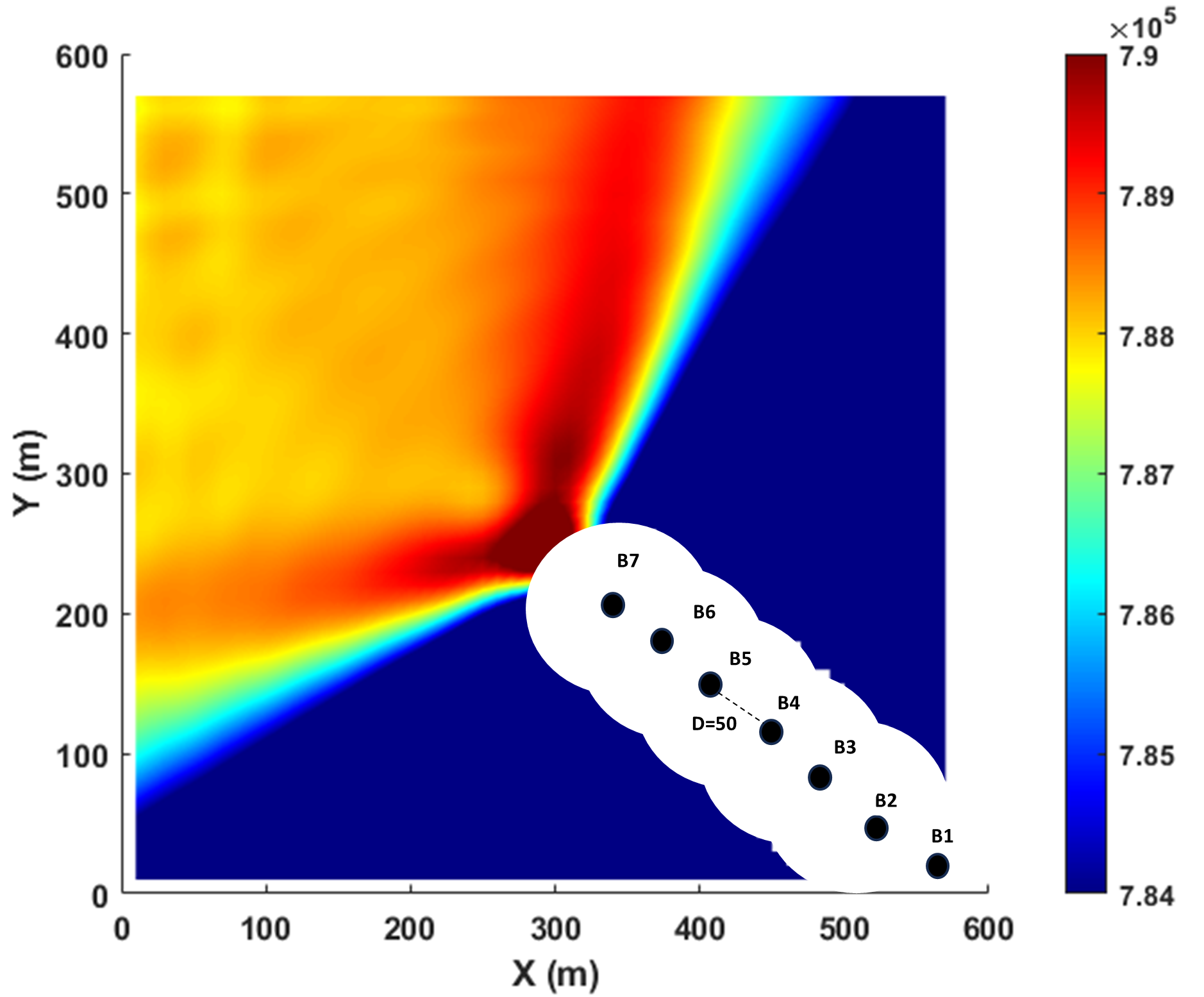}}
 \subfloat[]{
 \includegraphics[clip,width=0.25\columnwidth]{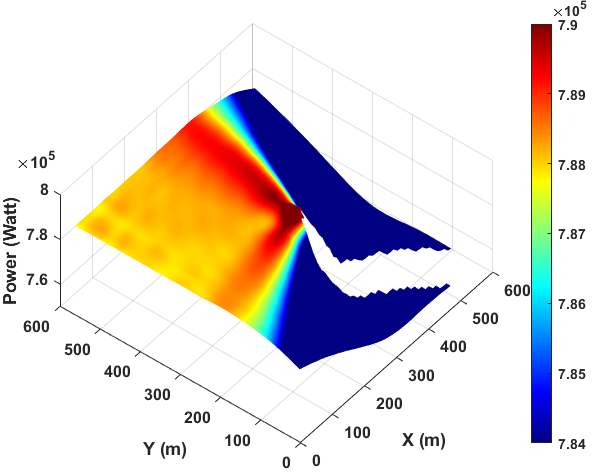}}\\
 \subfloat[]{
 \includegraphics[clip,width=0.25\columnwidth]{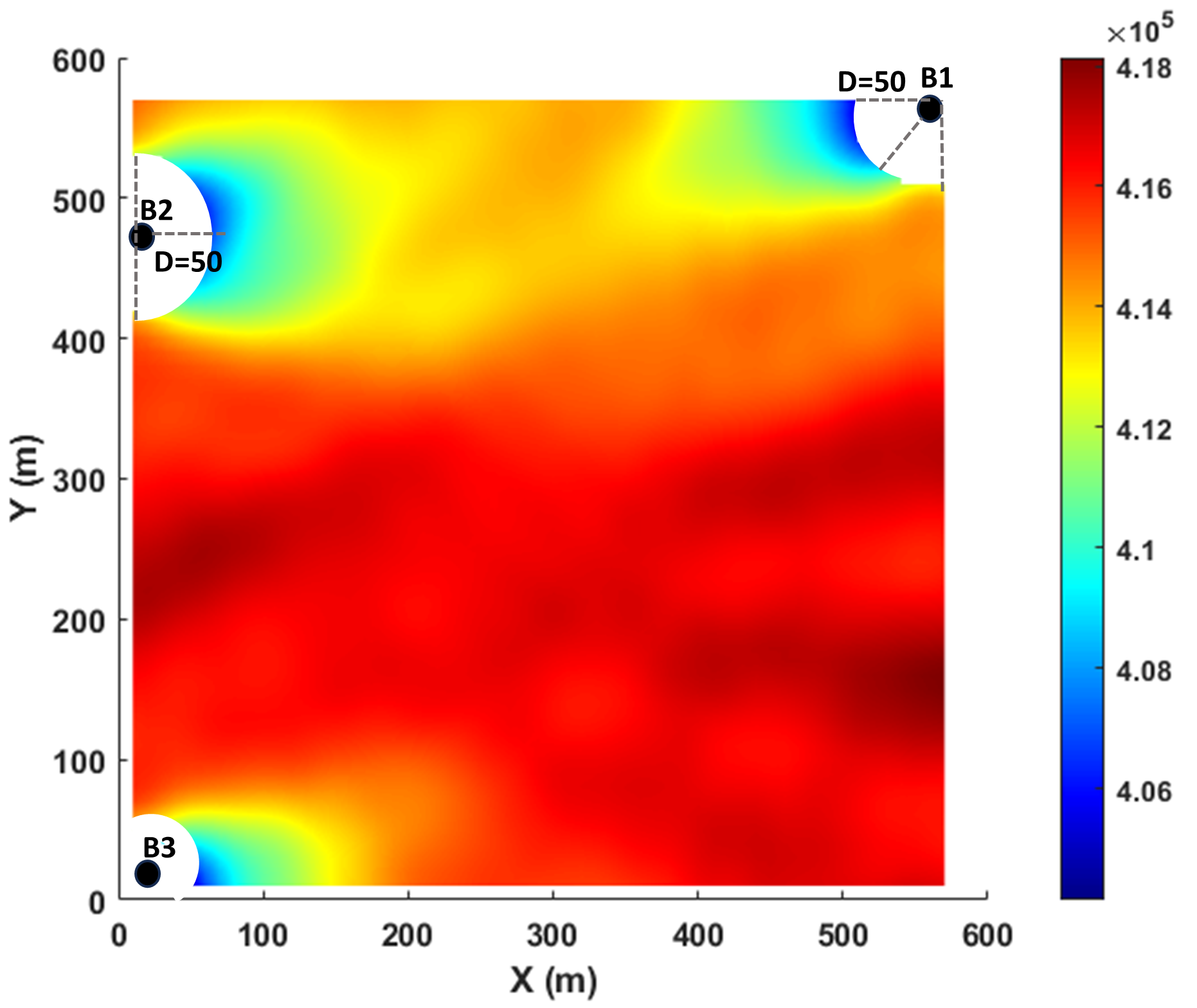}}
 \subfloat[]{
 \includegraphics[clip,width=0.25\columnwidth]{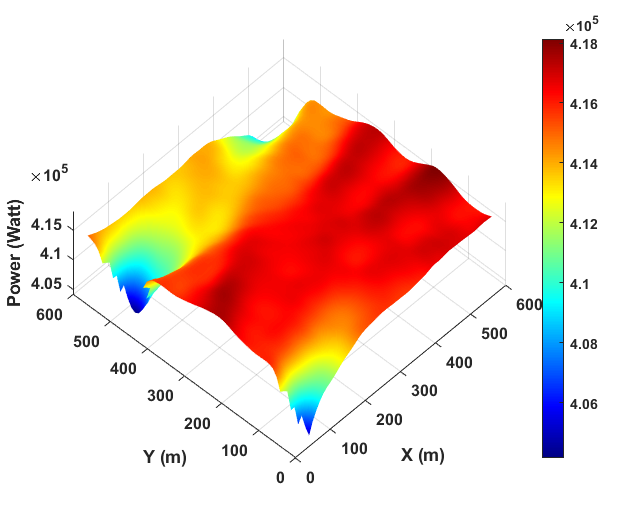}}
 \subfloat[]{
 \includegraphics[clip,width=0.25\columnwidth]{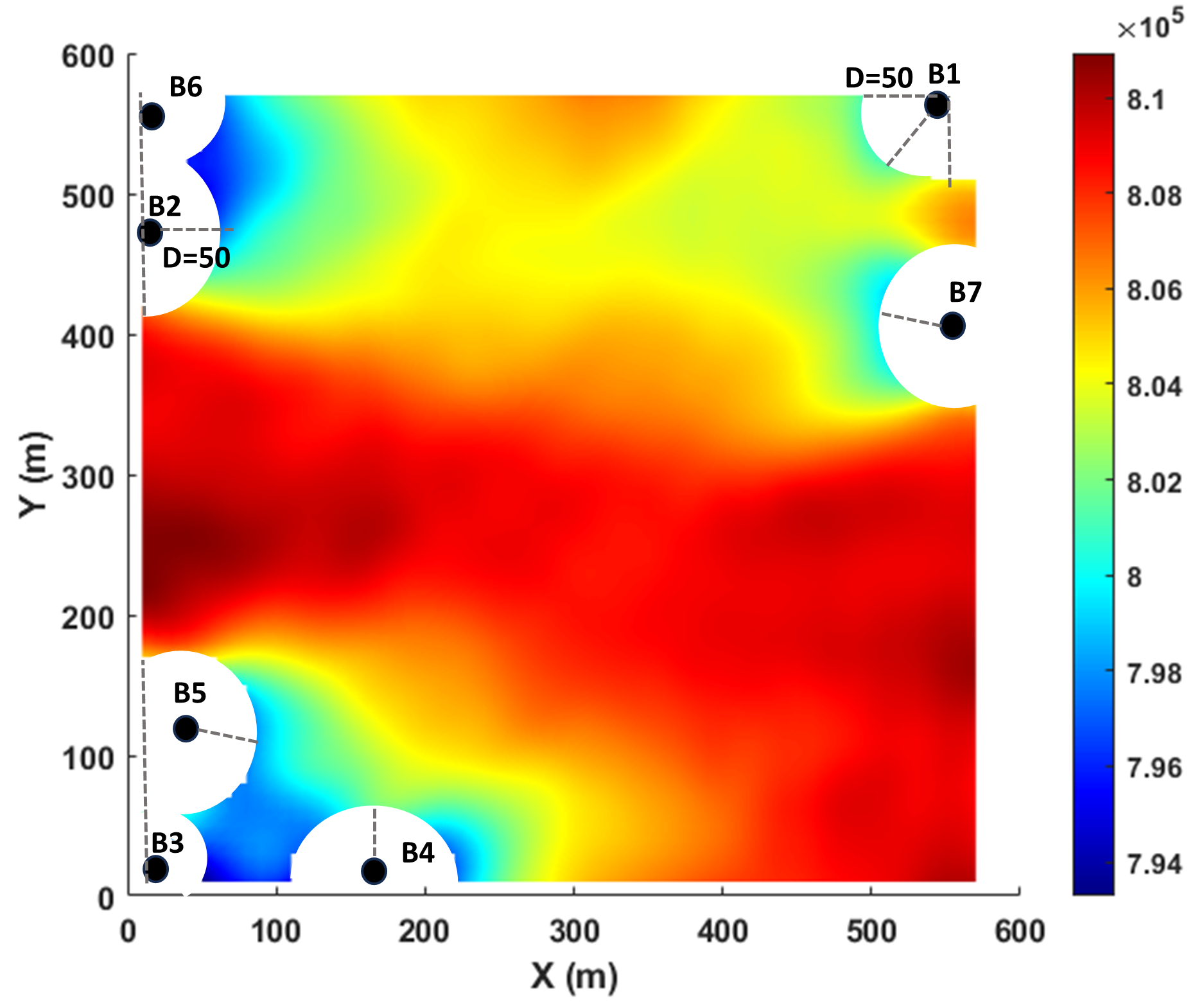}}
 \subfloat[]{
 \includegraphics[clip,width=0.25\columnwidth]{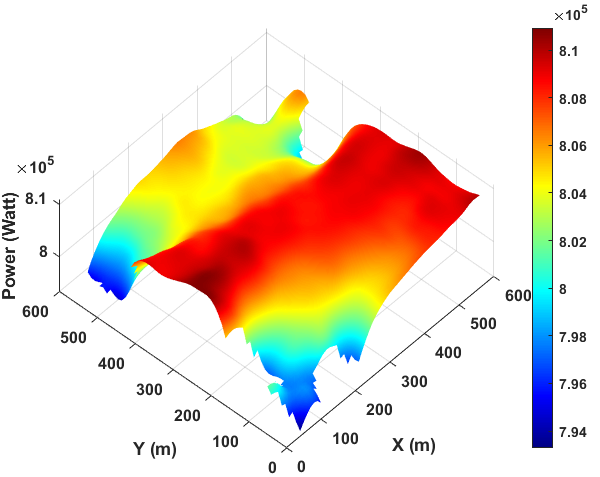}}
\caption{ Total power output Landscape analysis of the farm best-configuration based on Adelaide and Sydney sea site. The evaluation power of the fourth WEC through a whole feasible wave farm location with the initial three fixed WECs can be seen in (a and b) in the Adelaide site and (e and f) in the Sydney site. Various power potentials of the eighth WEC with the initial seven fixed WECs are shown by (c and d) in Adelaide and (g and h) based on the Sydney sea site. The black circles show the WECs with a safe distance equal to 50 m.   }
\label{fig:lanscape}
\end{figure}

The results of the experiments, depicted in Figure~\ref{fig:lanscape}, showcased the relationship between the total power output production and the relative positions of all buoys and wave interference in the wave farms. The analysis focused on two distinct wave climates, namely Adelaide and Sydney, providing insights into how the variations in wave characteristics in different locations can influence the power generation efficiency of the wave farms.
Interestingly, the power landscapes observed in the wave farms of Sydney and Adelaide exhibit distinct constructive and destructive wave interaction patterns. In the case of the four and eight WEC layouts in Adelaide, constructive wave interactions manifest as an unimodal region characterised by two edges that possess the highest potential for wave power extraction (see Figure~\ref{fig:lanscape} a-d). Besides, the absorbed power from a layout situated in Adelaide attains its peak value at a relatively short separation distance and a specific angle between the converters. These findings highlight the intricate relationship between wave interactions and power generation efficiency, emphasising the importance of carefully considering wave energy converters' spatial arrangement and orientation in optimising wave farm performance. The main difference between the analysis for the Adelaide and Sydney sites can be explained by differences in the directionality of the wave climates. The Adelaide site has narrowly spread seas, resulting in a clear directional preference for the potential installation of the next WEC for both array configurations. On the contrary, the Sydney site has wide directional spreading of the incoming waves, and there is no clear directional pattern for the last buoy installation.

It is interesting to acknowledge that the influence of separation distance on power generation varies depending on wave climates. For the Sydney site, an increase in the separation distance positively affects the power generation efficiency. This behaviour can be attributed to the complex nature of wave directions observed in Sydney, leading to a multifaceted power landscape depicted in Figure~\ref{fig:lanscape} (e-h). The power landscape exhibits a multimodal characteristic, shaped by the adjustments in the position of the last WEC in both the four and eight-WEC layouts. 

\subsection{Results and Finding}

In this study, nine distinct long-short-term memory ($LSTM$) models and a CNN were constructed, and their outputs were compared. Prior to the model's training, the number of input dimensions was adjusted to 32, and the total power value was predicted. Various parameter alterations were made to construct a model with a high degree of accuracy and to combat the overfitting problem.
The data was scaled and normalized using the Min-Max formula before training, which can be seen in the following.
\begin{equation}
x_{norm}^{i}=\frac{x^{i}-\mathbf{x}_{\min }}{\mathbf{x}_{\max }-\mathbf{x}_{\min }}
\end{equation}

We used eight evaluation metrics, including The Mean Squared Error (MSE), Root Mean Squared Error (RMSE), Loss function, Mean absolute error (MAE), Coefficient of Determination ($R^2$), Mean Squared Logarithmic Error (MSLE), Median Absolute Error (MedAE), and $Max_{Error}$ values of the training and test data were analyzed to assess the goodness-of-fit of the model. The technical details of these metrics can be seen in Table~\ref{tab:metrics}.

\begin{table}[h!]
\caption{Performance metrics and formulations.}
    \label{tab:metrics}  
    \centering
    \scalebox{0.85}{
\begin{tabular}{c|c}
\hline Parameters & Formula \\
\hline Mean Squared Error (MSE) & $M S E=\frac{1}{N} \sum_{i=1}^N\left(y_i-\hat{y}\right)^2$, Where,$\hat{y}$ is the predicted value of $y$ \\
\hline Root Mean Squared Error (RMSE) & $R M S E=\sqrt{\frac{1}{N} \sum_{i=1}^N\left(y_i-\hat{y}\right)^2}$ \\
\hline LOSS & Loss $=\frac{1}{2 N} \sum_{i=1}^N\left(y_i-\hat{y}\right)^2$ \\
\hline Mean absolute error (MAE) & $M A E=\frac{1}{N} \sum_{i=1}^N\left|y_i-\hat{y}\right|$ \\
\hline Coefficient of Determination ($R^2$) & $R^2=1-\frac{\sum\left(y_i-\hat{y}\right)^2}{\sum\left(y_i-\bar{y}\right)^2}$, Where, $\bar{y}$ is the mean value of $y$ \\
\hline Mean Squared Logarithmic Error (MSLE) & $M S L E=\frac{1}{N} \sum_{i=1}^N\left(\left(\log \left(1+y_i\right)\right)-(\log (1+\hat{y}))\right)^2$ \\
\hline Median Absolute Error (MedAE) & MedAE $=$ median $(y-\hat{y})$ \\
\hline $MAX_{ERROR}$ & $\operatorname{Max}_{Erorr}=\frac{1}{N} \sum_{i=1}^N\left|\hat{y}-y_i\right|$ \\
\hline
\end{tabular}
 }
\end{table}
Table~\ref{table:results_LSTM} presents the statistical results of two different LSTM models: a one-layer LSTM (Vanilla) model and a stacked model with two layers. To ensure a comprehensive evaluation, the average, minimum, maximum, and standard deviation of predicted power outputs were considered. These results were obtained through a 10-fold cross-validation method, enhancing the accuracy of the comparison.
An important observation is that increasing the number of LSTM layers does not guarantee superior performance in predicting wave farm power output across all wave scenarios. For instance, when examining the  Tasmania dataset, the stacked LSTM model failed to outperform the LSTM  model in terms of $R^2$, with the LSTM outperforming it by 22\%. However, in other wave farms, such as Adelaide, Perth, and Sydney, the stacked  LSTM model exhibited better performance than the LSTM model by 8.3\%,  0.3\%, and 8.65\%, respectively.

\begin{table}[h!]
\centering
\caption{ Statistical results of total power output prediction for Vanilla-LSTM and Stacked-LSTM models based on four wave scenario}
\label{table:results_LSTM}
    \scalebox{0.65}{
\begin{tabular}{l|lllllll|l|lllllll}
\hlineB{4}
\multicolumn{8}{c}{Stacked-LSTM   (2 layers)}                                        & \multicolumn{8}{c}{Vanilla-LSTM}                                                      \\ \hlineB{4}
\multicolumn{8}{c}{Adelaide}                                                         & \multicolumn{8}{c}{Adelaide}                                                         \\ \hlineB{1}
Metric & RMSE     & LOSS     & MAE      & R2       & MSLE     & MEDAE    & Max-error & Metric & RMSE     & LOSS     & MAE      & R2       & MSLE     & MEDAE    & Max-error \\ \hlineB{1}
Mean   & 1.16E-01 & 6.73E-03 & 9.27E-02 & 5.49E-01 & 5.82E-03 & 7.95E-02 & 4.92E-01  & Mean   & 1.20E-01 & 7.28E-03 & 9.64E-02 & 5.07E-01 & 6.28E-03 & 8.19E-02 & 5.02E-01  \\
Min    & 9.83E-02 & 4.84E-03 & 7.89E-02 & 5.23E-01 & 4.08E-03 & 6.76E-02 & 4.62E-01  & Min    & 1.02E-01 & 5.23E-03 & 8.15E-02 & 4.60E-01 & 4.40E-03 & 6.92E-02 & 4.62E-01  \\
Max    & 1.28E-01 & 8.14E-03 & 1.03E-01 & 5.88E-01 & 7.19E-03 & 8.88E-02 & 5.24E-01  & Max    & 1.32E-01 & 8.71E-03 & 1.05E-01 & 5.87E-01 & 7.72E-03 & 8.95E-02 & 5.40E-01  \\
STD    & 8.36E-03 & 9.49E-04 & 6.73E-03 & 2.30E-02 & 1.02E-03 & 6.28E-03 & 1.79E-02  & STD    & 8.00E-03 & 9.34E-04 & 6.56E-03 & 3.44E-02 & 9.98E-04 & 6.53E-03 & 2.49E-02  \\ \hlineB{1}
\multicolumn{8}{c}{Perth}                                                            & \multicolumn{8}{c}{Perth}                                                            \\\hlineB{1}
Metric & RMSE     & LOSS     & MAE      & R2       & MSLE     & MEDAE    & Max-error & Metric & RMSE     & LOSS     & MAE      & R2       & MSLE     & MEDAE    & Max-error \\\hlineB{1}
Mean   & 1.26E-01 & 7.91E-03 & 9.94E-02 & 3.51E-01 & 7.01E-03 & 8.30E-02 & 5.27E-01  & Mean   & 1.26E-01 & 7.97E-03 & 1.00E-01 & 3.50E-01 & 7.07E-03 & 8.41E-02 & 5.36E-01  \\
Min    & 1.16E-01 & 6.72E-03 & 9.13E-02 & 2.49E-01 & 5.38E-03 & 7.49E-02 & 4.96E-01  & Min    & 1.16E-01 & 6.77E-03 & 9.14E-02 & 3.05E-01 & 5.42E-03 & 7.44E-02 & 5.11E-01  \\
Max    & 1.42E-01 & 1.00E-02 & 1.13E-01 & 4.42E-01 & 9.25E-03 & 9.74E-02 & 5.62E-01  & Max    & 1.38E-01 & 9.57E-03 & 1.11E-01 & 3.90E-01 & 8.87E-03 & 9.44E-02 & 5.58E-01  \\
STD    & 7.65E-03 & 9.86E-04 & 6.64E-03 & 6.66E-02 & 1.08E-03 & 6.84E-03 & 2.08E-02  & STD    & 7.16E-03 & 9.08E-04 & 6.11E-03 & 3.27E-02 & 1.06E-03 & 6.30E-03 & 1.40E-02  \\\hlineB{1}
\multicolumn{8}{c}{Tasmania}                                                         & \multicolumn{8}{c}{Tasmania}                                                         \\\hlineB{1}
Metric & RMSE     & LOSS     & MAE      & R2       & MSLE     & MEDAE    & Max-error & Metric & RMSE     & LOSS     & MAE      & R2       & MSLE     & MEDAE    & Max-error \\\hlineB{1}
Mean   & 1.16E-01 & 6.70E-03 & 9.02E-02 & 2.05E-01 & 6.14E-03 & 7.43E-02 & 5.13E-01  & Mean   & 1.13E-01 & 6.40E-03 & 8.79E-02 & 2.50E-01 & 5.88E-03 & 7.21E-02 & 5.14E-01  \\
Min    & 1.03E-01 & 5.32E-03 & 8.12E-02 & 1.83E-01 & 4.55E-03 & 6.77E-02 & 4.91E-01  & Min    & 1.01E-01 & 5.05E-03 & 7.90E-02 & 2.25E-01 & 4.33E-03 & 6.43E-02 & 5.01E-01  \\
Max    & 1.28E-01 & 8.13E-03 & 9.96E-02 & 2.45E-01 & 7.90E-03 & 8.22E-02 & 5.38E-01  & Max    & 1.25E-01 & 7.83E-03 & 9.73E-02 & 2.70E-01 & 7.63E-03 & 8.04E-02 & 5.22E-01  \\
STD    & 7.45E-03 & 8.62E-04 & 5.89E-03 & 1.83E-02 & 1.03E-03 & 4.88E-03 & 1.46E-02  & STD    & 7.70E-03 & 8.66E-04 & 6.04E-03 & 1.62E-02 & 1.02E-03 & 4.92E-03 & 6.38E-03  \\\hlineB{1}
\multicolumn{8}{c}{Sydney}                                                           & \multicolumn{8}{c}{Sydney}                                                           \\\hlineB{1}
Metric & RMSE     & LOSS     & MAE      & R2       & MSLE     & MEDAE    & Max-error & Metric & RMSE     & LOSS     & MAE      & R2       & MSLE     & MEDAE    & Max-error \\\hlineB{1}
Mean   & 1.32E-01 & 8.75E-03 & 1.04E-01 & 2.89E-01 & 6.81E-03 & 8.75E-02 & 6.59E-01  & Mean   & 1.34E-01 & 8.98E-03 & 1.05E-01 & 2.66E-01 & 6.98E-03 & 9.01E-02 & 6.63E-01  \\
Min    & 1.15E-01 & 6.61E-03 & 9.24E-02 & 2.22E-01 & 4.74E-03 & 7.99E-02 & 6.14E-01  & Min    & 1.16E-01 & 6.70E-03 & 9.31E-02 & 2.23E-01 & 4.80E-03 & 8.25E-02 & 6.15E-01  \\
Max    & 1.51E-01 & 1.14E-02 & 1.17E-01 & 3.99E-01 & 9.33E-03 & 9.54E-02 & 7.11E-01  & Max    & 1.51E-01 & 1.14E-02 & 1.17E-01 & 3.19E-01 & 9.35E-03 & 9.59E-02 & 7.25E-01  \\
STD    & 1.18E-02 & 1.56E-03 & 8.43E-03 & 5.70E-02 & 1.46E-03 & 6.23E-03 & 3.05E-02  & STD    & 1.10E-02 & 1.46E-03 & 7.52E-03 & 3.17E-02 & 1.40E-03 & 4.09E-03 & 3.30E-02 \\\hlineB{4} 
\end{tabular}
}
\end{table}

One of the most popular sequential deep learning models is Bidirectional LSTM (BiLSTM) due to its ability to handle long-term dependencies better than traditional LSTM. By capturing information from both the past and future, BiLSTM provides an improved understanding of context. Therefore, we conducted a comparison between BiLSTM and its stacked version with two layers in predicting the power output at four sea sites. The results are presented in Table~\ref{table:results_BiLSTM}.
The stacked BiLSTM model outperformed the standard BiLSTM model among the sea sites examined, except for the Tasmania wave model. Specifically, in the Adelaide, Perth, and Sydney sea sites, the stacked  BiLSTM model performed better by 3.29\%, 4.10\%, and 10.38\%, respectively.

\begin{table}[h!]
\centering
\caption{ Statistical results of total power output prediction for Vanilla-GRU and Stacked-GRU models based on four wave scenario}
\label{table:results_GRU}
    \scalebox{0.65}{
\begin{tabular}{l|lllllll|l|lllllll}
\hlineB{4}
\multicolumn{8}{c}{Stacked-GRU}                                                      & \multicolumn{8}{c}{Vanilla-GRU}                                                      \\\hlineB{4}
\multicolumn{8}{c}{Adelaide}                                                         & \multicolumn{8}{c}{Adelaide}                                                         \\\hlineB{1}
Metric & RMSE     & LOSS     & MAE      & R2       & MSLE     & MEDAE    & Max-error & Metric & RMSE     & LOSS     & MAE      & R2       & MSLE     & MEDAE    & Max-error \\\hlineB{1}
Mean   & 1.12E-01 & 6.27E-03 & 8.94E-02 & 5.93E-01 & 5.46E-03 & 7.69E-02 & 4.76E-01  & Mean   & 1.23E-01 & 7.62E-03 & 9.87E-02 & 4.88E-01 & 6.55E-03 & 8.43E-02 & 5.13E-01  \\
Min    & 9.34E-02 & 4.36E-03 & 7.49E-02 & 5.80E-01 & 3.71E-03 & 6.49E-02 & 4.52E-01  & Min    & 1.11E-01 & 6.15E-03 & 8.84E-02 & 4.45E-01 & 5.12E-03 & 7.48E-02 & 4.88E-01  \\
Max    & 1.26E-01 & 7.92E-03 & 1.01E-01 & 6.12E-01 & 7.01E-03 & 8.72E-02 & 4.95E-01  & Max    & 1.36E-01 & 9.28E-03 & 1.09E-01 & 5.36E-01 & 8.22E-03 & 9.49E-02 & 5.53E-01  \\
STD    & 9.40E-03 & 1.04E-03 & 7.48E-03 & 9.90E-03 & 1.08E-03 & 6.63E-03 & 1.64E-02  & STD    & 8.42E-03 & 1.05E-03 & 6.92E-03 & 2.94E-02 & 1.11E-03 & 7.23E-03 & 1.82E-02  \\\hlineB{1}
\multicolumn{8}{c}{Perth}                                                            & \multicolumn{8}{c}{Perth}                                                            \\\hlineB{1}
Metric & RMSE     & LOSS     & MAE      & R2       & MSLE     & MEDAE    & Max-error & Metric & RMSE     & LOSS     & MAE      & R2       & MSLE     & MEDAE    & Max-error \\ \hlineB{1}
Mean   & 1.18E-01 & 6.99E-03 & 9.33E-02 & 4.64E-01 & 6.27E-03 & 7.84E-02 & 5.09E-01  & Mean   & 1.24E-01 & 7.72E-03 & 9.84E-02 & 3.82E-01 & 6.87E-03 & 8.25E-02 & 5.26E-01  \\
Min    & 1.09E-01 & 5.99E-03 & 8.65E-02 & 3.96E-01 & 4.83E-03 & 7.07E-02 & 4.81E-01  & Min    & 1.15E-01 & 6.64E-03 & 9.02E-02 & 3.15E-01 & 5.31E-03 & 7.30E-02 & 5.03E-01  \\
Max    & 1.29E-01 & 8.28E-03 & 1.03E-01 & 5.40E-01 & 7.79E-03 & 8.85E-02 & 5.34E-01  & Max    & 1.34E-01 & 9.01E-03 & 1.07E-01 & 4.38E-01 & 8.40E-03 & 9.31E-02 & 5.44E-01  \\
STD    & 6.67E-03 & 7.96E-04 & 5.61E-03 & 4.07E-02 & 9.31E-04 & 5.80E-03 & 1.51E-02  & STD    & 6.38E-03 & 7.97E-04 & 5.56E-03 & 4.07E-02 & 9.49E-04 & 6.16E-03 & 1.33E-02  \\\hlineB{1}
\multicolumn{8}{c}{Tasmania}                                                         & \multicolumn{8}{c}{Tasmania}                                                         \\\hlineB{1}
Metric & RMSE     & LOSS     & MAE      & R2       & MSLE     & MEDAE    & Max-error & Metric & RMSE     & LOSS     & MAE      & R2       & MSLE     & MEDAE    & Max-error \\\hlineB{1}
Mean   & 1.13E-01 & 6.44E-03 & 8.81E-02 & 2.47E-01 & 5.92E-03 & 7.20E-02 & 5.06E-01  & Mean   & 1.12E-01 & 6.24E-03 & 8.67E-02 & 2.75E-01 & 5.74E-03 & 7.09E-02 & 5.03E-01  \\
Min    & 9.99E-02 & 4.99E-03 & 7.82E-02 & 1.57E-01 & 4.27E-03 & 6.45E-02 & 4.73E-01  & Min    & 9.96E-02 & 4.96E-03 & 7.82E-02 & 2.59E-01 & 4.25E-03 & 6.41E-02 & 4.91E-01  \\
Max    & 1.29E-01 & 8.32E-03 & 1.01E-01 & 2.94E-01 & 8.06E-03 & 8.36E-02 & 5.39E-01  & Max    & 1.24E-01 & 7.64E-03 & 9.62E-02 & 3.02E-01 & 7.45E-03 & 7.90E-02 & 5.22E-01  \\
STD    & 8.33E-03 & 9.53E-04 & 6.68E-03 & 3.74E-02 & 1.10E-03 & 5.61E-03 & 2.11E-02  & STD    & 7.39E-03 & 8.24E-04 & 5.82E-03 & 1.52E-02 & 9.89E-04 & 4.85E-03 & 8.34E-03  \\\hlineB{1}
\multicolumn{8}{c}{Sydney}                                                           & \multicolumn{8}{c}{Sydney}                                                           \\\hlineB{1}
Metric & RMSE     & LOSS     & MAE      & R2       & MSLE     & MEDAE    & Max-error & Metric & RMSE     & LOSS     & MAE      & R2       & MSLE     & MEDAE    & Max-error \\\hlineB{1}
Mean   & 1.28E-01 & 8.34E-03 & 1.00E-01 & 3.45E-01 & 6.53E-03 & 8.41E-02 & 6.41E-01  & Mean   & 1.32E-01 & 8.73E-03 & 1.04E-01 & 2.96E-01 & 6.81E-03 & 8.85E-02 & 6.60E-01  \\
Min    & 1.08E-01 & 5.82E-03 & 8.66E-02 & 1.35E-01 & 4.20E-03 & 7.58E-02 & 6.00E-01  & Min    & 1.16E-01 & 6.68E-03 & 9.21E-02 & 2.70E-01 & 4.79E-03 & 8.04E-02 & 6.21E-01  \\
Max    & 1.63E-01 & 1.32E-02 & 1.27E-01 & 4.13E-01 & 1.07E-02 & 1.08E-01 & 6.81E-01  & Max    & 1.47E-01 & 1.08E-02 & 1.13E-01 & 3.39E-01 & 8.86E-03 & 9.50E-02 & 7.12E-01  \\
STD    & 1.56E-02 & 2.10E-03 & 1.17E-02 & 8.26E-02 & 1.85E-03 & 9.14E-03 & 2.89E-02  & STD    & 1.14E-02 & 1.48E-03 & 7.99E-03 & 2.47E-02 & 1.41E-03 & 4.79E-03 & 2.80E-02 \\\hlineB{4}
\end{tabular}
}
\end{table}
\begin{table}[h!]
\centering
\caption{ Statistical results of total power output prediction for CNN and CNN-LSTM models based on four wave scenario}
\label{table:results_CNN}
    \scalebox{0.65}{
\begin{tabular}{l|lllllll|l|lllllll}
\hlineB{4}
\multicolumn{8}{c}{CNN}                                                              & \multicolumn{8}{c}{CNN-LSTM}                                                         \\ \hlineB{4}
\multicolumn{8}{c}{Adelaide}                                                         & \multicolumn{8}{c}{Adelaide}                                                         \\\hlineB{1}
Metric & RMSE     & LOSS     & MAE      & R2       & MSLE     & MEDAE    & Max-error & Metric & RMSE     & LOSS     & MAE      & R2       & MSLE     & MEDAE    & Max-error \\\hlineB{1}
Mean   & 8.83E-02 & 3.93E-03 & 6.72E-02 & 6.97E-01 & 3.59E-03 & 5.16E-02 & 4.61E-01  & Mean   & 7.44E-02 & 2.78E-03 & 5.48E-02 & 7.97E-01 & 2.68E-03 & 4.03E-02 & 4.31E-01  \\
Min    & 7.60E-02 & 2.89E-03 & 5.64E-02 & 6.46E-01 & 2.64E-03 & 4.05E-02 & 4.19E-01  & Min    & 6.63E-02 & 2.20E-03 & 4.76E-02 & 7.80E-01 & 2.08E-03 & 3.34E-02 & 3.58E-01  \\
Max    & 1.01E-01 & 5.13E-03 & 7.76E-02 & 7.38E-01 & 4.97E-03 & 6.34E-02 & 5.31E-01  & Max    & 8.20E-02 & 3.36E-03 & 6.02E-02 & 8.28E-01 & 3.35E-03 & 5.07E-02 & 4.96E-01  \\
STD    & 8.27E-03 & 7.33E-04 & 7.07E-03 & 3.22E-02 & 7.19E-04 & 7.46E-03 & 3.70E-02  & STD    & 5.18E-03 & 3.86E-04 & 4.62E-03 & 1.35E-02 & 4.25E-04 & 5.82E-03 & 4.21E-02  \\\hlineB{1}
\multicolumn{8}{c}{Perth}                                                            & \multicolumn{8}{c}{Perth}                                                            \\\hlineB{1}
Metric & RMSE     & LOSS     & MAE      & R2       & MSLE     & MEDAE    & Max-error & Metric & RMSE     & LOSS     & MAE      & R2       & MSLE     & MEDAE    & Max-error \\\hlineB{1}
Mean   & 8.83E-02 & 3.90E-03 & 6.46E-02 & 6.59E-01 & 3.71E-03 & 4.66E-02 & 4.95E-01  & Mean   & 7.45E-02 & 2.78E-03 & 5.46E-02 & 7.62E-01 & 2.67E-03 & 3.94E-02 & 4.20E-01  \\
Min    & 8.36E-02 & 3.49E-03 & 6.06E-02 & 6.25E-01 & 3.18E-03 & 4.26E-02 & 4.05E-01  & Min    & 7.03E-02 & 2.47E-03 & 5.33E-02 & 7.53E-01 & 2.24E-03 & 3.80E-02 & 3.82E-01  \\
Max    & 9.67E-02 & 4.68E-03 & 7.08E-02 & 7.06E-01 & 4.57E-03 & 5.11E-02 & 5.94E-01  & Max    & 7.90E-02 & 3.12E-03 & 5.80E-02 & 7.73E-01 & 3.05E-03 & 4.22E-02 & 4.69E-01  \\
STD    & 4.41E-03 & 3.93E-04 & 3.22E-03 & 2.76E-02 & 4.21E-04 & 2.65E-03 & 5.62E-02  & STD    & 2.63E-03 & 1.97E-04 & 1.76E-03 & 5.90E-03 & 2.32E-04 & 1.46E-03 & 3.12E-02  \\\hlineB{1}
\multicolumn{8}{c}{Tasmania}                                                         & \multicolumn{8}{c}{Tasmania}                                                         \\\hlineB{1}
Metric & RMSE     & LOSS     & MAE      & R2       & MSLE     & MEDAE    & Max-error & Metric & RMSE     & LOSS     & MAE      & R2       & MSLE     & MEDAE    & Max-error \\\hlineB{1}
Mean   & 8.63E-02 & 3.74E-03 & 6.56E-02 & 5.21E-01 & 3.54E-03 & 5.05E-02 & 4.66E-01  & Mean   & 7.46E-02 & 2.80E-03 & 5.69E-02 & 6.73E-01 & 2.71E-03 & 4.42E-02 & 3.75E-01  \\
Min    & 7.82E-02 & 3.05E-03 & 5.96E-02 & 4.67E-01 & 2.81E-03 & 4.56E-02 & 3.63E-01  & Min    & 6.70E-02 & 2.25E-03 & 5.11E-02 & 6.59E-01 & 2.05E-03 & 3.87E-02 & 3.35E-01  \\
Max    & 1.01E-01 & 5.14E-03 & 7.67E-02 & 5.80E-01 & 5.09E-03 & 5.91E-02 & 5.45E-01  & Max    & 8.68E-02 & 3.76E-03 & 6.62E-02 & 7.04E-01 & 3.78E-03 & 5.17E-02 & 4.19E-01  \\
STD    & 6.41E-03 & 5.77E-04 & 4.89E-03 & 3.54E-02 & 6.25E-04 & 4.37E-03 & 5.81E-02  & STD    & 6.92E-03 & 5.33E-04 & 5.27E-03 & 1.28E-02 & 5.96E-04 & 4.25E-03 & 3.18E-02  \\\hlineB{1}
\multicolumn{8}{c}{Sydney}                                                           & \multicolumn{8}{c}{Sydney}                                                           \\\hlineB{1}
Metric & RMSE     & LOSS     & MAE      & R2       & MSLE     & MEDAE    & Max-error & Metric & RMSE     & LOSS     & MAE      & R2       & MSLE     & MEDAE    & Max-error \\\hlineB{1}
Mean   & 8.31E-02 & 3.50E-03 & 5.87E-02 & 7.28E-01 & 2.80E-03 & 3.95E-02 & 5.58E-01  & Mean   & 6.40E-02 & 2.12E-03 & 4.59E-02 & 8.50E-01 & 1.72E-03 & 3.22E-02 & 4.00E-01  \\
Min    & 6.46E-02 & 2.08E-03 & 4.53E-02 & 6.81E-01 & 1.63E-03 & 2.99E-02 & 4.19E-01  & Min    & 5.23E-02 & 1.37E-03 & 3.83E-02 & 8.45E-01 & 1.04E-03 & 2.61E-02 & 3.06E-01  \\
Max    & 9.38E-02 & 4.40E-03 & 6.64E-02 & 7.87E-01 & 3.60E-03 & 4.57E-02 & 9.04E-01  & Max    & 8.30E-02 & 3.44E-03 & 6.01E-02 & 8.54E-01 & 2.89E-03 & 4.27E-02 & 4.74E-01  \\
STD    & 9.72E-03 & 7.83E-04 & 7.20E-03 & 3.19E-02 & 6.94E-04 & 5.49E-03 & 1.35E-01  & STD    & 1.21E-02 & 8.09E-04 & 8.78E-03 & 3.09E-03 & 7.18E-04 & 6.49E-03 & 6.06E-02 \\ \hlineB{4}
\end{tabular}
}
\end{table}
In the evaluation of wave farm power prediction models, the Gated  Recurrent Unit (GRU) was tested using the following machine learning model. GRU  incorporates a simplification of the LSTM architecture by combining the input and forgets gates into a single update gate. This modification reduces computational complexity and enhances the training speed of GRU  models. The statistical analysis presented in Table~\ref{table:results_GRU} compares the performance of both GRU and stacked versions' performance using eight metrics. While the stacked model demonstrated greater accuracy in terms of the coefficient of determination ($R^2$), interestingly, similar to the LSTM  and Bi-LSTM models, a standalone GRU model produced more precise results compared to the stacked GRU model in the Tasmania wave model.

\begin{table}[h!]
\centering
\caption{ Statistical results of total power output prediction for CNN-BiLSTM and CNN-GRU models based on four wave scenario}
\label{table:results_CNN_Bilstm}
    \scalebox{0.65}{
\begin{tabular}{l|lllllll|l|lllllll}
\hlineB{4}
\multicolumn{8}{c}{CNN-BiLSTM}                                                       & \multicolumn{8}{c}{CNN-GRU}                                                          \\ \hlineB{1}
\multicolumn{8}{c}{Adelaide}                                                         & \multicolumn{8}{c}{Adelaide}                                                         \\ \hlineB{1}
Metric & RMSE     & LOSS     & MAE      & R2       & MSLE     & MEDAE    & Max-error & Metric & RMSE     & LOSS     & MAE      & R2       & MSLE     & MEDAE    & Max-error \\ \hlineB{1}
Mean   & 7.12E-02 & 2.56E-03 & 5.27E-02 & 8.12E-01 & 2.46E-03 & 3.89E-02 & 4.03E-01  & Mean   & 7.26E-02 & 2.66E-03 & 5.36E-02 & 7.84E-01 & 2.50E-03 & 3.91E-02 & 4.36E-01  \\
Min    & 5.71E-02 & 1.63E-03 & 4.18E-02 & 7.85E-01 & 1.54E-03 & 3.03E-02 & 3.45E-01  & Min    & 6.71E-02 & 2.25E-03 & 4.87E-02 & 7.76E-01 & 2.10E-03 & 3.42E-02 & 3.61E-01  \\
Max    & 8.15E-02 & 3.32E-03 & 6.07E-02 & 8.57E-01 & 3.35E-03 & 5.10E-02 & 5.06E-01  & Max    & 8.70E-02 & 3.78E-03 & 6.39E-02 & 7.96E-01 & 3.78E-03 & 5.07E-02 & 5.00E-01  \\
STD    & 8.56E-03 & 6.02E-04 & 7.04E-03 & 1.86E-02 & 6.35E-04 & 7.10E-03 & 4.87E-02  & STD    & 6.67E-03 & 5.08E-04 & 5.57E-03 & 6.04E-03 & 5.60E-04 & 5.97E-03 & 3.97E-02  \\ \hlineB{1}
\multicolumn{8}{c}{Perth}                                                            & \multicolumn{8}{c}{Perth}                                                            \\\hlineB{1}
Metric & RMSE     & LOSS     & MAE      & R2       & MSLE     & MEDAE    & Max-error & Metric & RMSE     & LOSS     & MAE      & R2       & MSLE     & MEDAE    & Max-error \\\hlineB{1}
Mean   & 7.50E-02 & 2.83E-03 & 5.58E-02 & 7.85E-01 & 2.73E-03 & 4.16E-02 & 3.96E-01  & Mean   & 7.60E-02 & 2.89E-03 & 5.64E-02 & 7.51E-01 & 2.75E-03 & 4.14E-02 & 4.06E-01  \\
Min    & 6.53E-02 & 2.13E-03 & 4.86E-02 & 7.62E-01 & 2.04E-03 & 3.59E-02 & 3.24E-01  & Min    & 7.18E-02 & 2.57E-03 & 5.36E-02 & 7.36E-01 & 2.32E-03 & 3.78E-02 & 3.72E-01  \\
Max    & 8.57E-02 & 3.68E-03 & 6.54E-02 & 8.26E-01 & 3.64E-03 & 5.15E-02 & 4.49E-01  & Max    & 8.16E-02 & 3.33E-03 & 5.98E-02 & 7.60E-01 & 3.25E-03 & 4.39E-02 & 4.43E-01  \\
STD    & 6.05E-03 & 4.59E-04 & 4.85E-03 & 2.29E-02 & 5.09E-04 & 4.77E-03 & 3.33E-02  & STD    & 3.72E-03 & 2.85E-04 & 2.02E-03 & 8.63E-03 & 3.66E-04 & 1.84E-03 & 2.15E-02  \\\hlineB{1}
\multicolumn{8}{c}{Tasmania}                                                         & \multicolumn{8}{c}{Tasmania}                                                         \\\hlineB{1}
Metric & RMSE     & LOSS     & MAE      & R2       & MSLE     & MEDAE    & Max-error & Metric & RMSE     & LOSS     & MAE      & R2       & MSLE     & MEDAE    & Max-error \\\hlineB{1}
Mean   & 7.19E-02 & 2.61E-03 & 5.46E-02 & 6.88E-01 & 2.52E-03 & 4.22E-02 & 3.75E-01  & Mean   & 7.54E-02 & 2.85E-03 & 5.75E-02 & 6.52E-01 & 2.73E-03 & 4.47E-02 & 3.94E-01  \\
Min    & 6.25E-02 & 1.95E-03 & 4.84E-02 & 6.59E-01 & 1.79E-03 & 3.83E-02 & 3.27E-01  & Min    & 6.69E-02 & 2.24E-03 & 5.16E-02 & 6.17E-01 & 2.04E-03 & 4.03E-02 & 3.64E-01  \\
Max    & 8.54E-02 & 3.64E-03 & 6.49E-02 & 7.08E-01 & 3.67E-03 & 5.06E-02 & 4.28E-01  & Max    & 8.48E-02 & 3.59E-03 & 6.40E-02 & 6.69E-01 & 3.58E-03 & 4.87E-02 & 4.36E-01  \\
STD    & 7.66E-03 & 5.63E-04 & 5.37E-03 & 1.42E-02 & 6.31E-04 & 3.99E-03 & 3.24E-02  & STD    & 5.62E-03 & 4.30E-04 & 3.87E-03 & 1.57E-02 & 5.07E-04 & 3.12E-03 & 2.45E-02  \\\hlineB{1}
\multicolumn{8}{c}{Sydney}                                                           & \multicolumn{8}{c}{Sydney}                                                           \\\hlineB{1}
Metric & RMSE     & LOSS     & MAE      & R2       & MSLE     & MEDAE    & Max-error & Metric & RMSE     & LOSS     & MAE      & R2       & MSLE     & MEDAE    & Max-error \\\hlineB{1}
Mean   & 6.26E-02 & 2.02E-03 & 4.49E-02 & 8.55E-01 & 1.63E-03 & 3.17E-02 & 3.80E-01  & Mean   & 6.55E-02 & 2.20E-03 & 4.69E-02 & 8.37E-01 & 1.77E-03 & 3.28E-02 & 4.23E-01  \\
Min    & 5.24E-02 & 1.38E-03 & 3.74E-02 & 8.49E-01 & 1.05E-03 & 2.58E-02 & 3.17E-01  & Min    & 5.45E-02 & 1.48E-03 & 3.99E-02 & 8.24E-01 & 1.13E-03 & 2.84E-02 & 3.23E-01  \\
Max    & 8.12E-02 & 3.30E-03 & 6.00E-02 & 8.63E-01 & 2.75E-03 & 4.50E-02 & 4.74E-01  & Max    & 8.61E-02 & 3.71E-03 & 6.24E-02 & 8.48E-01 & 3.11E-03 & 4.47E-02 & 5.14E-01  \\
STD    & 1.15E-02 & 7.56E-04 & 8.67E-03 & 4.57E-03 & 6.67E-04 & 6.93E-03 & 6.06E-02  & STD    & 1.13E-02 & 7.86E-04 & 8.03E-03 & 6.97E-03 & 7.05E-04 & 5.83E-03 & 7.35E-02 \\\hlineB{1}
\end{tabular}
}
\end{table}
Upon conducting tests and comparisons of LSTM, Bi-LSTM, and GRU models alongside their stacked versions, our focus shifted towards  Convolutional deep learning models. These models employ convolutional layers and pooling operations, enabling them to effectively capture spatial or temporal patterns inherent in the data. To enhance the capabilities of CNNs, we devised a hybrid model that integrates both CNN and LSTM architectures. The rationale behind this approach stems from the fact that combining CNNs with LSTMs allows for further processing and extraction of hierarchical representations acquired by the CNN  layers, transforming them into higher-level features through the LSTM  layers. Consequently, this amalgamation enables the model to leverage the local pattern recognition abilities of CNNs as well as the sequential modelling capabilities of LSTMs. 
\begin{figure}[h!]
\centering
\subfloat[]{
 \includegraphics[clip,width=0.5\columnwidth]{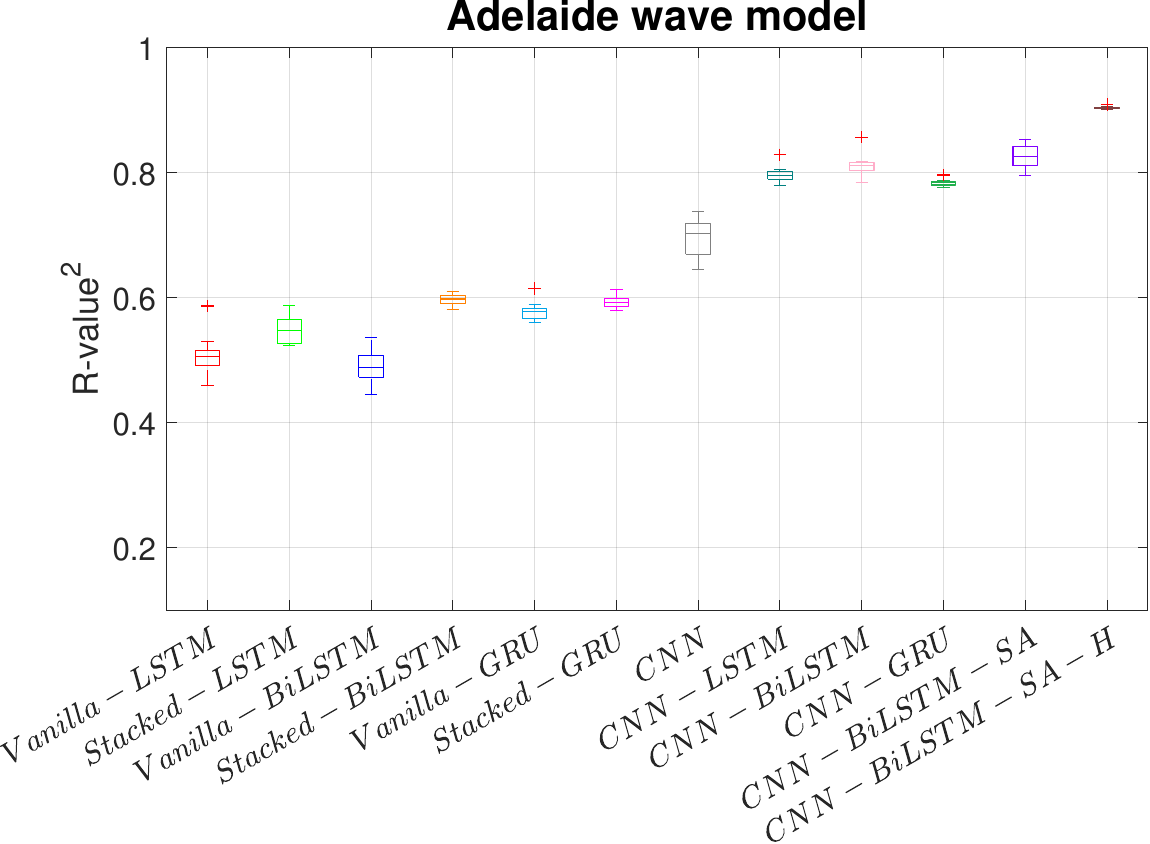}}
 \subfloat[]{
 \includegraphics[clip,width=0.5\columnwidth]{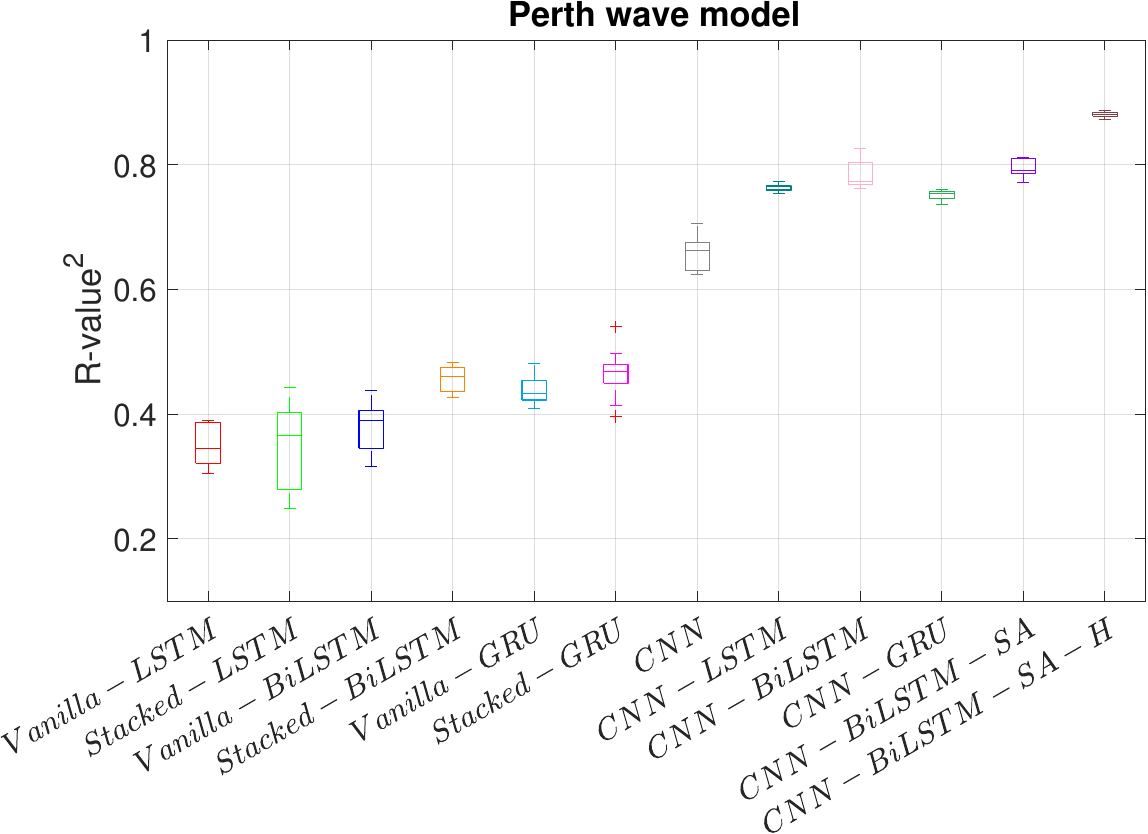}}\\
 \subfloat[]{
 \includegraphics[clip,width=0.5\columnwidth]{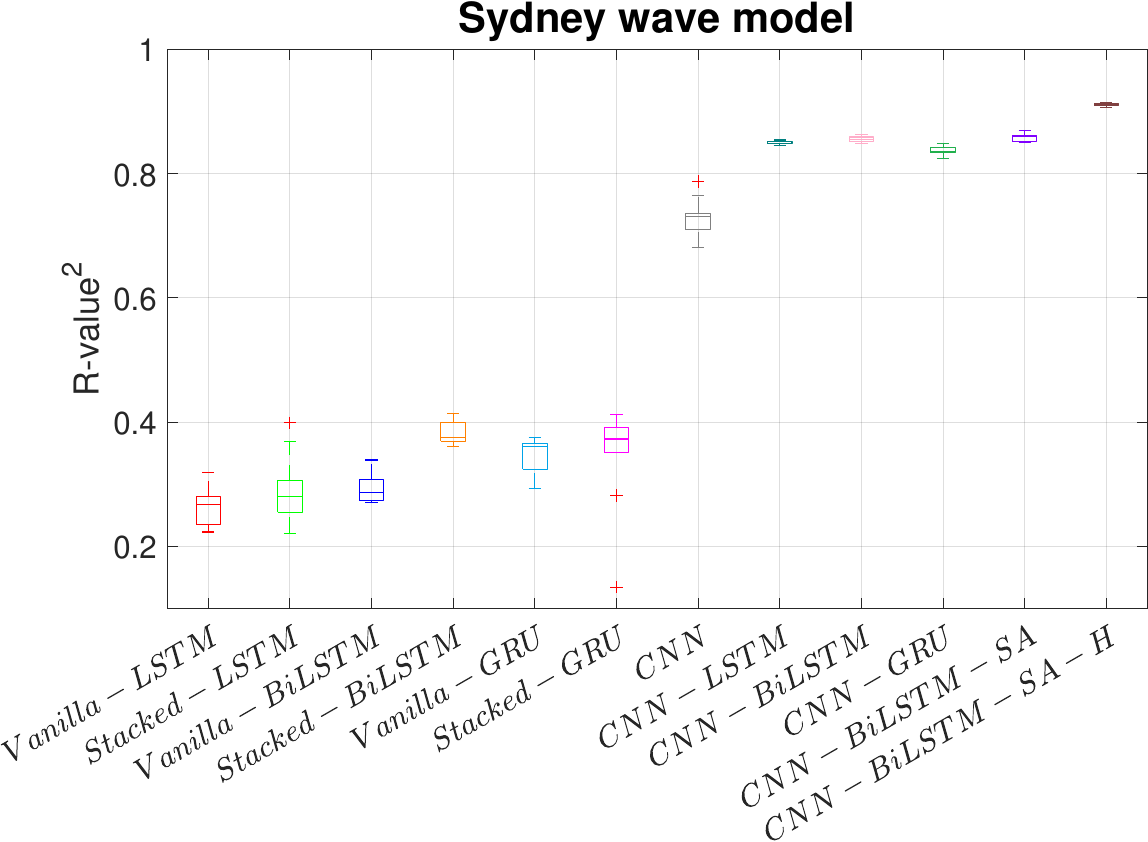}}
 \subfloat[]{
 \includegraphics[clip,width=0.5\columnwidth]{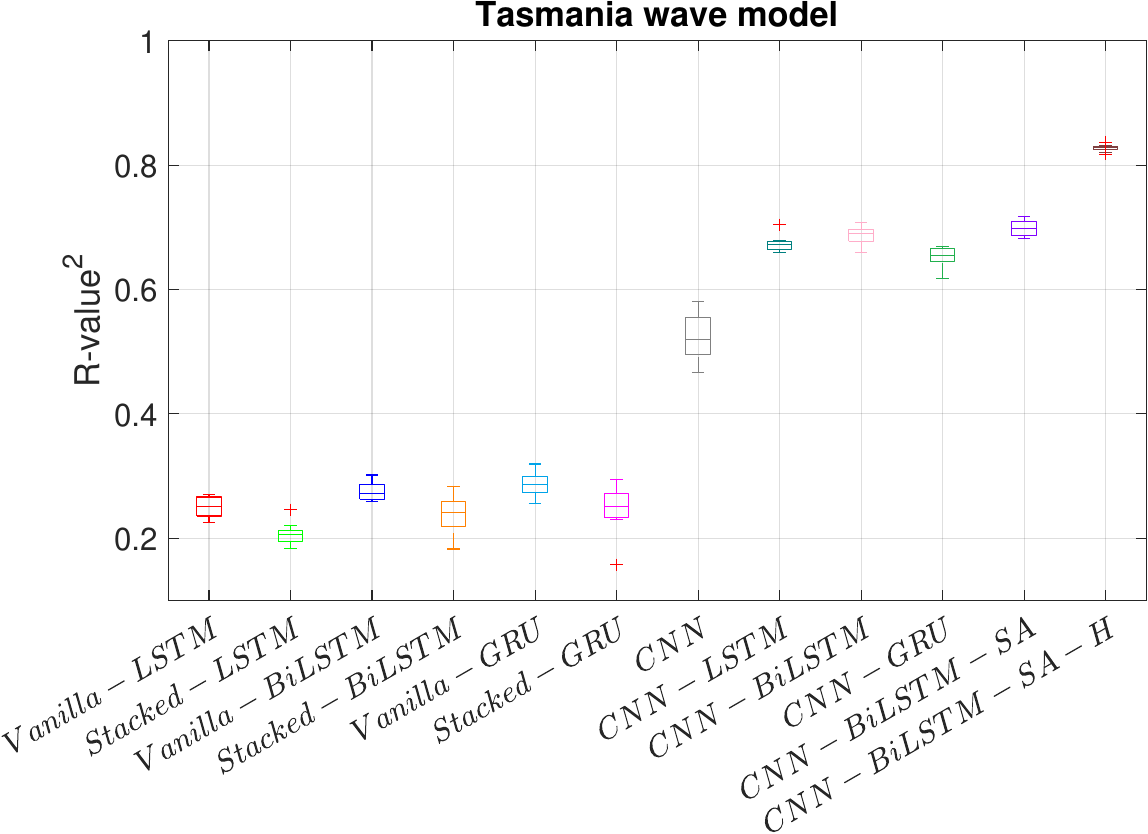}}
  \caption{ Statistical details of wave power prediction models comparison based on $R^2$ (accuracy) for four-wave models, (a) Adelaide, (b) Perth, (c) Sydney, (d) Tasmania sea sites.   }
\label{fig:prediction_models_all}
\end{figure}
Table~\ref{table:results_CNN} presents the prediction outcomes of both the CNN and the hybrid  CNN-LSTM model for four distinct wave case studies. Notably, the CNN  exhibits a remarkable capability to discern the complex and nonlinear relationship between the position of the WECs and the power generated by the wave farm. Consequently, the CNN model delivers more precise and robust results compared to all previously tested models. Moreover, incorporating LSTM layers significantly enhances the performance of the CNN, as demonstrated in Table~\ref{table:results_CNN}. The CNN-LSTM model surpasses the CNN model by margins of 14.4\%, 15.62\%, 29.18\%, and 16.80\% in the respective locations of Adelaide, Perth,  Tasmania, and Sydney in terms of $R^2$.

This study evaluated the performance of three hybrid models: CNN-LSTM, CNN-BiLSTM, and CNN-GRU. Among these models,  the highest level of performance was observed with the CNN-BiLSTM architecture across four distinct sea sites, as demonstrated in Table~\ref{table:results_CNN_Bilstm}. The CNN-BiLSTM model exhibited exceptional accuracy in predicting the power output, achieving impressive percentages of 85.5\%, 81.2\%, 78.5\%,  and 68.8\% in Sydney, Adelaide, Perth, and Tasmania, respectively. These results highlight the superior predictive capabilities of the CNN-BiLSTM hybrid model, positioning it as a promising choice for accurate power output predictions in the context of wave energy applications.
\begin{table}[h!]
\centering
\caption{ Statistical results of total power output prediction for CNN-BiLSTM-SA and CNN-BiLSTM-SA-E models based on four wave scenario}
\label{table:results_CNN_Bilstm_SA_H}
    \scalebox{0.65}{
\begin{tabular}{l|lllllll|l|lllllll}
\hlineB{4}
\multicolumn{8}{c}{CNN-BiLSTM-SA}                                                    & \multicolumn{8}{c}{CNN-BiLSTM-SA-H}                        \\\hlineB{4}
\multicolumn{8}{c}{Adelaide}                                                         & \multicolumn{8}{c}{Adelaide}                               \\\hlineB{1}
Metric & RMSE     & LOSS     & MAE      & R2       & MSLE     & MEDAE    & Max-error & Metric & RMSE      & LOSS      & MAE       & R2        & MSLE      & MEDAE     & Max-error \\ \hlineB{1}
Mean   & 6.99E-02 & 2.47E-03 & 5.21E-02 & 8.26E-01 & 2.36E-03 & 3.92E-02 & 4.03E-01  & Mean   & 4.521E-02 & 1.023E-03 & 3.506E-02 & 9.041E-01 & 9.175E-04 & 2.837E-02 & 2.626E-01 \\
Min    & 5.75E-02 & 1.65E-03 & 4.24E-02 & 7.96E-01 & 1.57E-03 & 3.07E-02 & 3.51E-01  & Min    & 4.294E-02 & 9.219E-04 & 3.292E-02 & 9.016E-01 & 8.307E-04 & 2.583E-02 & 2.289E-01 \\
Max    & 8.00E-02 & 3.20E-03 & 6.23E-02 & 8.53E-01 & 3.20E-03 & 5.72E-02 & 4.65E-01  & Max    & 4.641E-02 & 1.077E-03 & 3.600E-02 & 9.087E-01 & 9.752E-04 & 2.994E-02 & 2.934E-01 \\
STD    & 7.85E-03 & 5.44E-04 & 6.35E-03 & 1.89E-02 & 5.91E-04 & 7.63E-03 & 3.55E-02  & STD    & 9.317E-04 & 4.154E-05 & 1.004E-03 & 2.198E-03 & 3.663E-05 & 1.465E-03 & 2.124E-02 \\\hlineB{4}
       & \multicolumn{7}{c}{Perth}                                                   & \multicolumn{8}{c}{Perth}                                                                  \\\hlineB{4}
Metric & RMSE     & LOSS     & MAE      & R2       & MSLE     & MEDAE    & Max-error & Metric & RMSE      & LOSS      & MAE       & R2        & MSLE      & MEDAE     & Max-error \\\hlineB{1}
Mean   & 7.06E-02 & 2.51E-03 & 5.31E-02 & 7.93E-01 & 2.38E-03 & 3.98E-02 & 3.87E-01  & Mean   & 4.776E-02 & 1.142E-03 & 3.715E-02 & 8.808E-01 & 1.024E-03 & 3.022E-02 & 2.839E-01 \\
Min    & 6.27E-02 & 1.97E-03 & 4.85E-02 & 7.72E-01 & 1.77E-03 & 3.54E-02 & 3.39E-01  & Min    & 4.595E-02 & 1.056E-03 & 3.577E-02 & 8.724E-01 & 9.376E-04 & 2.906E-02 & 2.246E-01 \\
Max    & 8.67E-02 & 3.76E-03 & 6.61E-02 & 8.12E-01 & 3.75E-03 & 5.13E-02 & 4.21E-01  & Max    & 5.029E-02 & 1.264E-03 & 3.844E-02 & 8.867E-01 & 1.186E-03 & 3.193E-02 & 3.403E-01 \\
STD    & 6.67E-03 & 5.01E-04 & 4.93E-03 & 1.43E-02 & 5.72E-04 & 4.36E-03 & 2.99E-02  & STD    & 1.477E-03 & 7.081E-05 & 1.092E-03 & 4.594E-03 & 8.431E-05 & 1.156E-03 & 3.594E-02 \\\hlineB{4}
       & \multicolumn{7}{c}{Tasmania}                                                & \multicolumn{8}{c}{Tasmania}                                                               \\\hlineB{4}
Metric & RMSE     & LOSS     & MAE      & R2       & MSLE     & MEDAE    & Max-error & Metric & RMSE      & LOSS      & MAE       & R2        & MSLE      & MEDAE     & Max-error \\\hlineB{1}
Mean   & 6.83E-02 & 2.35E-03 & 5.22E-02 & 6.98E-01 & 2.25E-03 & 4.07E-02 & 3.52E-01  & Mean   & 4.698E-02 & 1.104E-03 & 3.693E-02 & 8.280E-01 & 1.001E-03 & 3.052E-02 & 2.539E-01 \\
Min    & 6.20E-02 & 1.92E-03 & 4.77E-02 & 6.82E-01 & 1.77E-03 & 3.73E-02 & 3.02E-01  & Min    & 4.532E-02 & 1.027E-03 & 3.560E-02 & 8.177E-01 & 9.313E-04 & 2.902E-02 & 2.095E-01 \\
Max    & 8.09E-02 & 3.27E-03 & 6.14E-02 & 7.17E-01 & 3.26E-03 & 4.77E-02 & 3.87E-01  & Max    & 4.897E-02 & 1.199E-03 & 3.918E-02 & 8.372E-01 & 1.078E-03 & 3.362E-02 & 3.418E-01 \\
STD    & 6.31E-03 & 4.49E-04 & 4.52E-03 & 1.26E-02 & 5.04E-04 & 3.51E-03 & 3.13E-02  & STD    & 1.045E-03 & 4.926E-05 & 1.098E-03 & 5.505E-03 & 4.183E-05 & 1.435E-03 & 4.133E-02 \\\hlineB{4}
       & \multicolumn{7}{c}{Sydney}                                                  & \multicolumn{8}{c}{Sydney}                                                                 \\\hlineB{4}
Metric & RMSE     & LOSS     & MAE      & R2       & MSLE     & MEDAE    & Max-error & Metric & RMSE      & LOSS      & MAE       & R2        & MSLE      & MEDAE     & Max-error \\\hlineB{1}
Mean   & 6.39E-02 & 2.09E-03 & 4.59E-02 & 8.59E-01 & 1.71E-03 & 3.24E-02 & 3.88E-01  & Mean   & 4.066E-02 & 8.273E-04 & 3.063E-02 & 9.105E-01 & 6.223E-04 & 2.355E-02 & 2.451E-01 \\
Min    & 4.89E-02 & 1.19E-03 & 3.60E-02 & 8.51E-01 & 9.05E-04 & 2.48E-02 & 2.72E-01  & Min    & 3.939E-02 & 7.757E-04 & 2.935E-02 & 9.067E-01 & 5.844E-04 & 2.116E-02 & 1.958E-01 \\
Max    & 8.20E-02 & 3.37E-03 & 6.04E-02 & 8.70E-01 & 2.85E-03 & 4.50E-02 & 4.72E-01  & Max    & 4.320E-02 & 9.330E-04 & 3.381E-02 & 9.140E-01 & 6.921E-04 & 2.843E-02 & 2.954E-01 \\
STD    & 1.11E-02 & 7.27E-04 & 8.29E-03 & 5.98E-03 & 6.55E-04 & 6.63E-03 & 6.38E-02  & STD    & 1.389E-03 & 5.750E-05 & 1.547E-03 & 2.568E-03 & 3.936E-05 & 2.270E-03 & 3.383E-02
   \\  \hlineB{4}
\end{tabular}
}
\end{table}
\begin{table}[h!]
\centering
\caption{ Optimal hyper-parameters configurations of CNN-BiLSTM-SA-H models based on four wave scenario}
\label{table:hyper-prameters}
    \scalebox{0.8}{\begin{tabular}{llll|ll|ll|l|l|l|l|l|l}
    \hlineB{4}  
\multicolumn{14}{c}{Tasmania}           \\ \hlineB{4}  
$CNF_1$ & $CNF_2$ & $CNF_3$ & $CNF_4$ & $NHU_1$ & $NHU_2$ & $PDO_1$     & $PDO_2$     & $BS$   & $LR$       & $AT$ & $L2Reg$    & $RMSE$  & $R^2$    \\  \hlineB{1}  
512  & 256  & 128  & 64   & 64   & 32   & 5.33E-02 & 5.33E-02 & 256  & 2.11E-03 & 16 & 1.03E-04 & 0.049 & 0.821 \\
512  & 256  & 128  & 64   & 16   & 8    & 6.93E-02 & 6.93E-02 & 1024 & 6.80E-04 & 64 & 1.18E-04 & 0.049 & 0.819 \\
512  & 256  & 128  & 64   & 64   & 32   & 5.62E-02 & 5.62E-02 & 512  & 5.32E-04 & 32 & 1.19E-04 & 0.049 & 0.821 \\
256  & 128  & 64   & 32   & 32   & 16   & 5.00E-02 & 5.00E-02 & 32   & 1.00E-04 & 32 & 1.00E-04 & 0.047 & 0.834 \\ \hlineB{4}  
\multicolumn{14}{c}{Perth}                                                                                      \\ \hlineB{4}  
$CNF_1$ & $CNF_2$ & $CNF_3$ & $CNF_4$ & $NHU_1$ & $NHU_2$ & $PDO_1$     & $PDO_2$     & $BS$   & $LR$       & $AT$ & $L2Reg$    & $RMSE$  & $R^2$     \\\hlineB{1}  
256  & 128  & 64   & 32   & 32   & 16   & 5.00E-02 & 5.00E-02 & 32   & 1.00E-04 & 32 & 1.00E-04 & 0.047 & 0.880 \\
256  & 128  & 64   & 32   & 8    & 4    & 5.46E-02 & 5.46E-02 & 128  & 2.00E-04 & 16 & 1.11E-04 & 0.046 & 0.882 \\
512  & 256  & 128  & 64   & 32   & 16   & 7.05E-02 & 7.05E-02 & 256  & 1.43E-04 & 64 & 1.09E-04 & 0.046 & 0.881 \\
512  & 256  & 128  & 64   & 8    & 4    & 5.52E-02 & 5.52E-02 & 512  & 2.33E-04 & 80 & 1.35E-04 & 0.046 & 0.881 \\\hlineB{4}  
\multicolumn{14}{c}{Adelaide}                                                                                   \\\hlineB{4}  
$CNF_1$ & $CNF_2$ & $CNF_3$ & $CNF_4$ & $NHU_1$ & $NHU_2$ & $PDO_1$     & $PDO_2$     & $BS$   & $LR$       & $AT$ & $L2Reg$    & $RMSE$  & $R^2$    \\\hlineB{1}  
256  & 128  & 64   & 32   & 32   & 16   & 5.00E-02 & 5.00E-02 & 32   & 1.00E-04 & 32 & 1.00E-04 & 0.044 & 0.905 \\
256  & 128  & 64   & 32   & 16   & 8    & 5.34E-02 & 5.34E-02 & 128  & 8.11E-04 & 32 & 1.05E-04 & 0.045 & 0.907 \\
512  & 256  & 128  & 64   & 8    & 4    & 7.19E-02 & 7.19E-02 & 128  & 4.89E-05 & 16 & 1.05E-04 & 0.044 & 0.909 \\
512  & 256  & 128  & 64   & 64   & 32   & 1.96E-01 & 1.96E-01 & 128  & 5.69E-05 & 80 & 1.01E-04 & 0.044 & 0.909 \\\hlineB{4}  
\multicolumn{14}{c}{Sydney}                                                                                     \\\hlineB{4}  
$CNF_1$ & $CNF_2$ & $CNF_3$ & $CNF_4$ & $NHU_1$ & $NHU_2$ & $PDO_1$     & $PDO_2$     & $BS$   & $LR$       & $AT$ & $L2Reg$    & $RMSE$  & $R^2$    \\\hlineB{1}  
256  & 128  & 64   & 32   & 32   & 16   & 5.00E-02 & 5.00E-02 & 32   & 1.00E-04 & 32 & 1.00E-04 & 0.040 & 0.911 \\
128  & 64   & 32   & 16   & 16   & 8    & 1.58E-01 & 1.58E-01 & 64   & 1.11E-04 & 64 & 2.71E-04 & 0.042 & 0.899 \\
256  & 128  & 64   & 32   & 64   & 32   & 8.24E-02 & 8.24E-02 & 64   & 5.89E-05 & 16 & 3.28E-04 & 0.041 & 0.906 \\
512  & 256  & 128  & 64   & 128  & 64   & 2.08E-01 & 2.08E-01 & 64   & 2.83E-05 & 48 & 1.23E-04 & 0.041 & 0.904 \\ \hlineB{4}  
\end{tabular}
}
\end{table}

\begin{figure}[h!]
\centering
\subfloat[]{
 \includegraphics[clip,width=0.5\columnwidth]{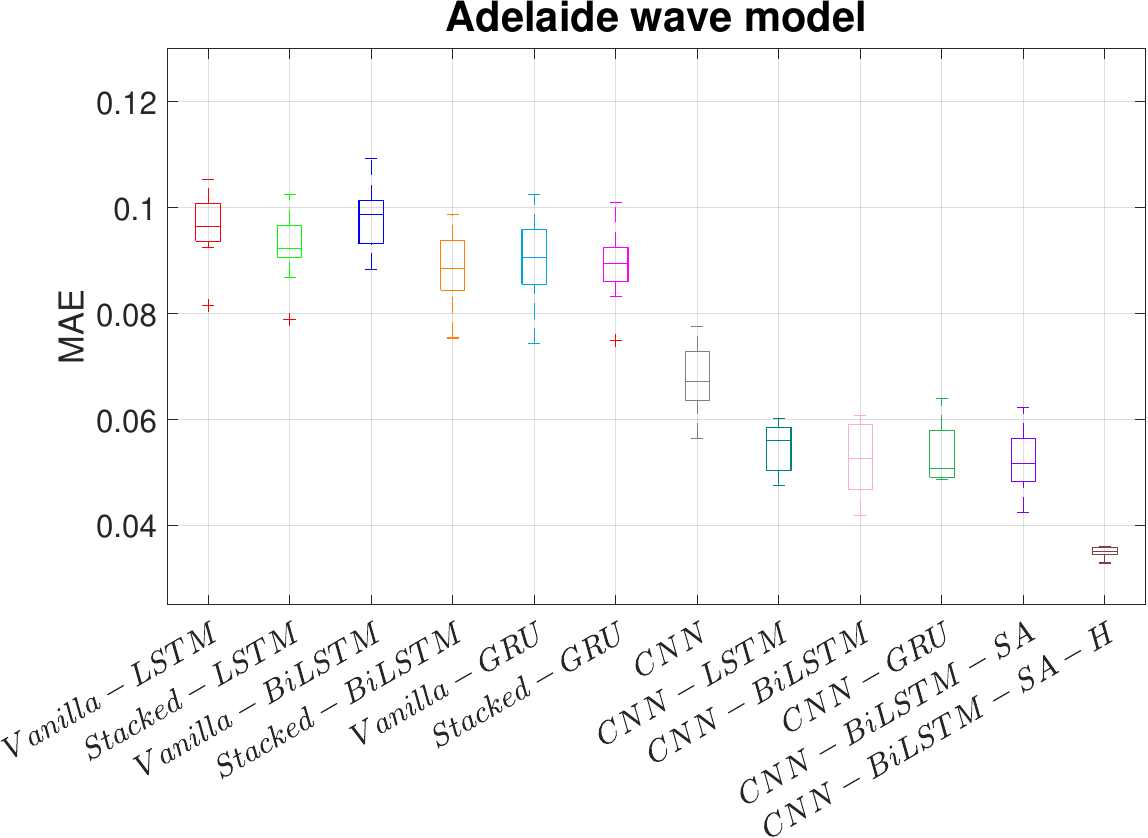}}
 \subfloat[]{
 \includegraphics[clip,width=0.5\columnwidth]{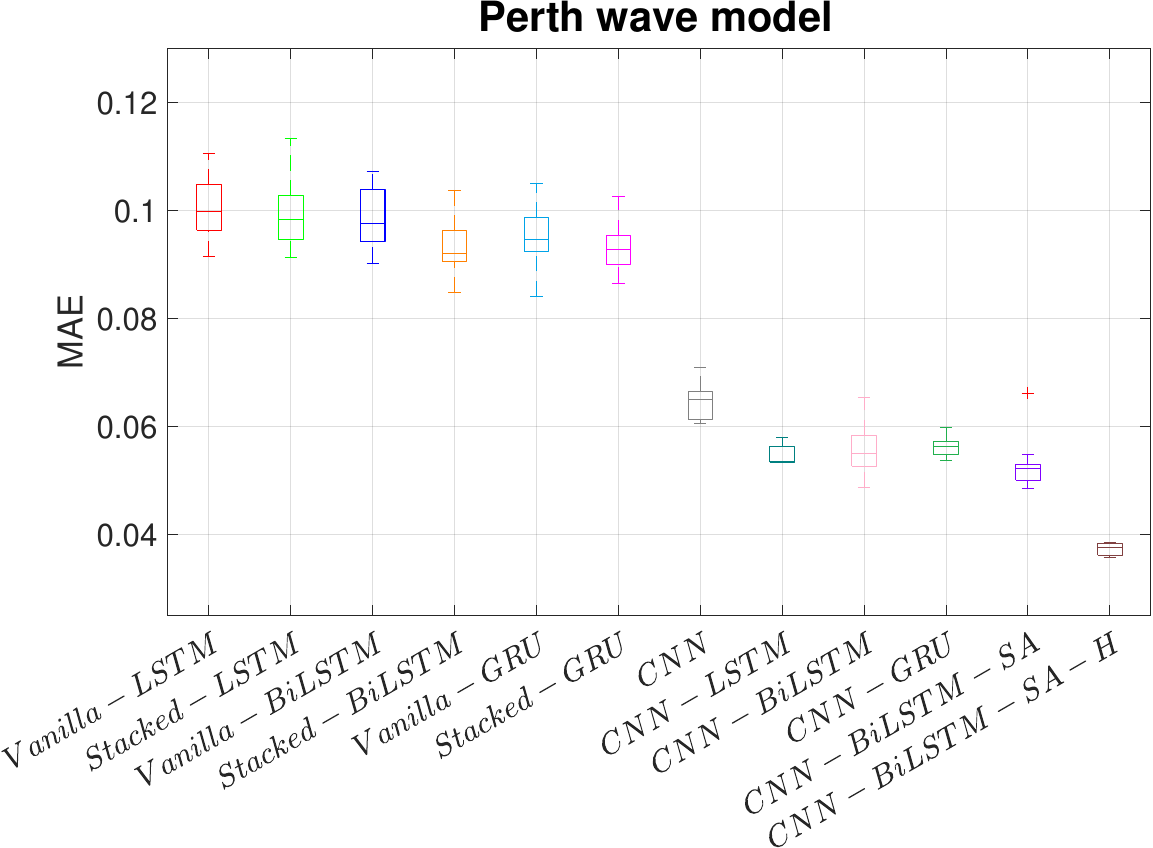}}\\
 \subfloat[]{
 \includegraphics[clip,width=0.5\columnwidth]{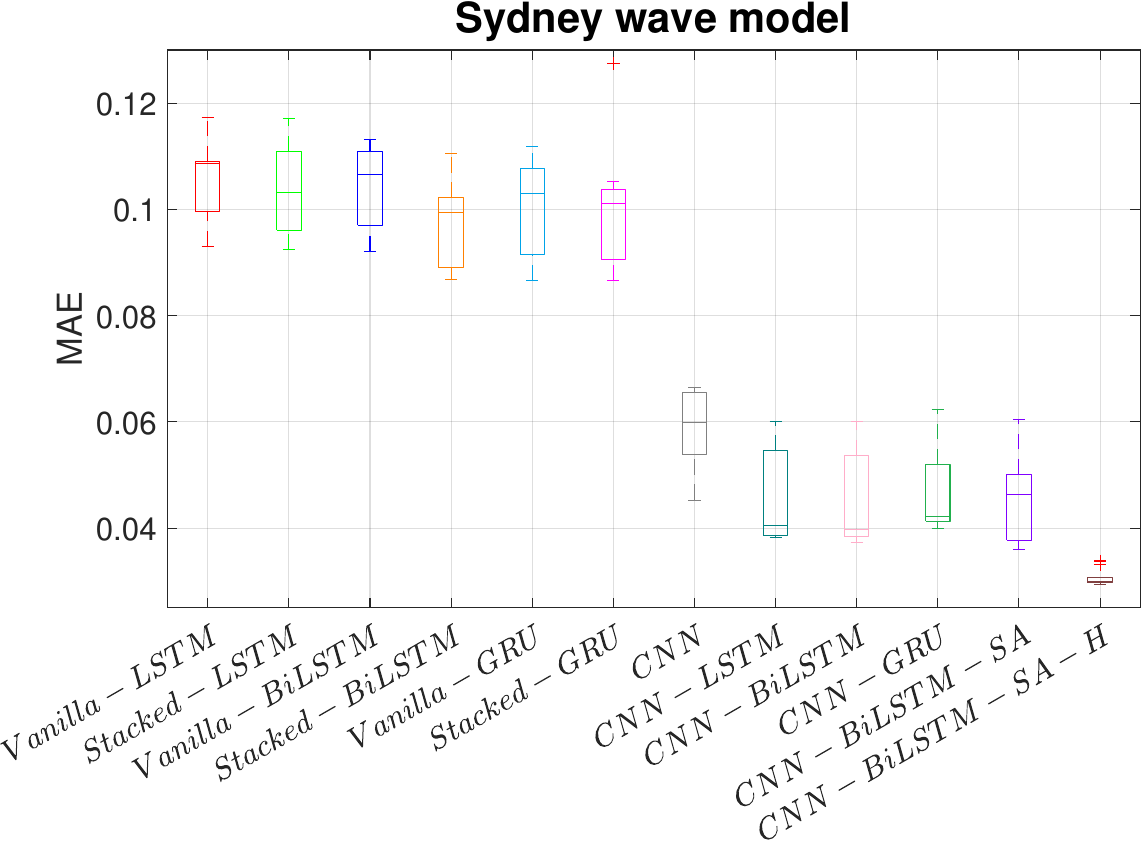}}
 \subfloat[]{
 \includegraphics[clip,width=0.5\columnwidth]{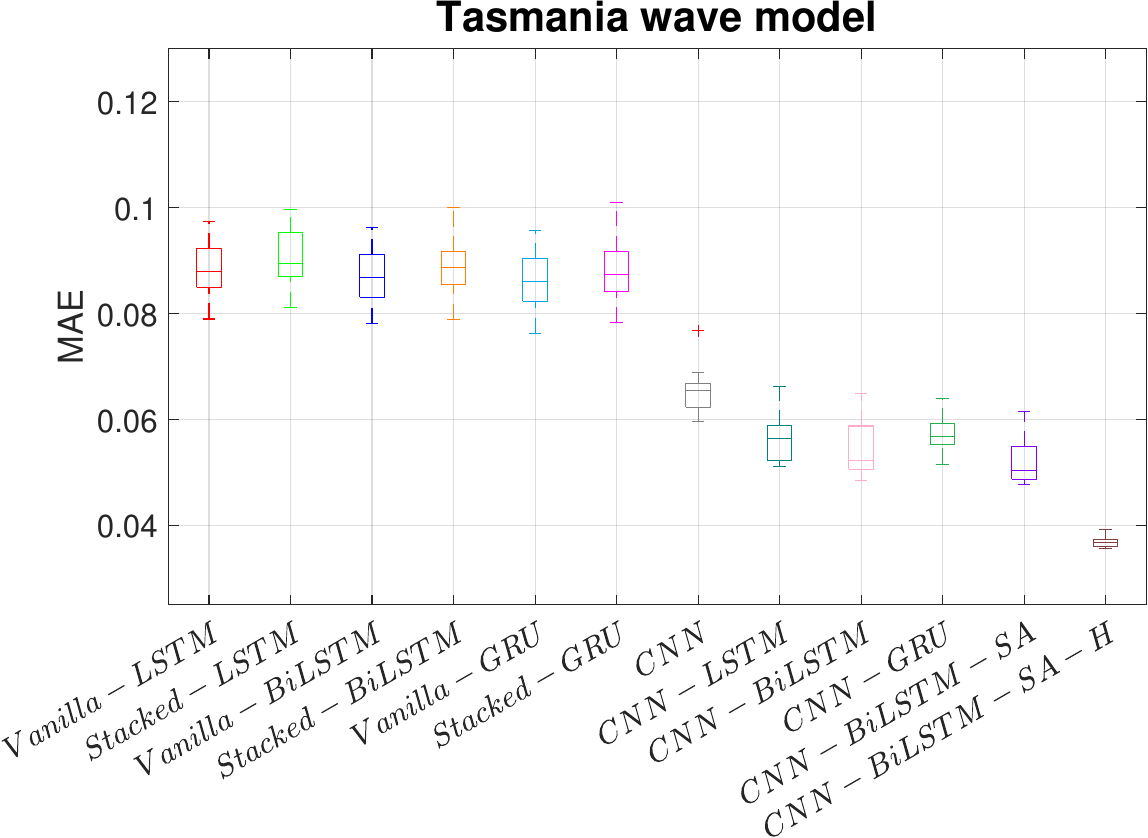}}
  \caption{Statistical details of wave power prediction models comparison based on MAE for four-wave models, (a) Adelaide, (b) Perth, (c) Sydney, (d) Tasmania sea sites.  }
\label{fig:prediction_mae_all}
\end{figure}
\begin{figure}[h!]
\centering
\subfloat[]{
 \includegraphics[clip,width=0.25\columnwidth]{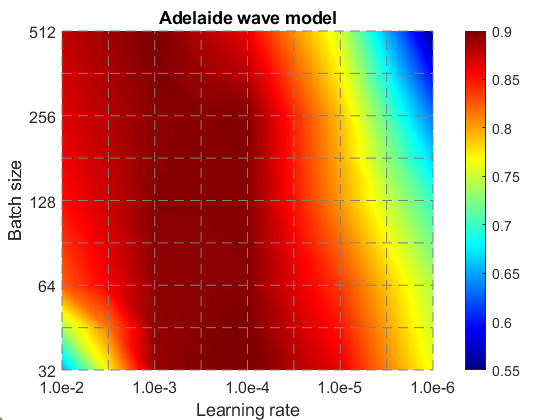}}
 \subfloat[]{
 \includegraphics[clip,width=0.25\columnwidth]{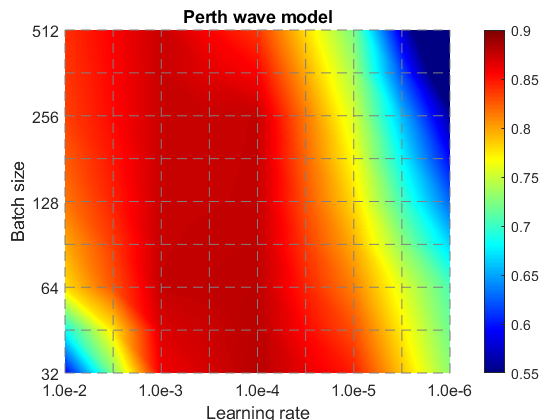}}
 \subfloat[]{
 \includegraphics[clip,width=0.25\columnwidth]{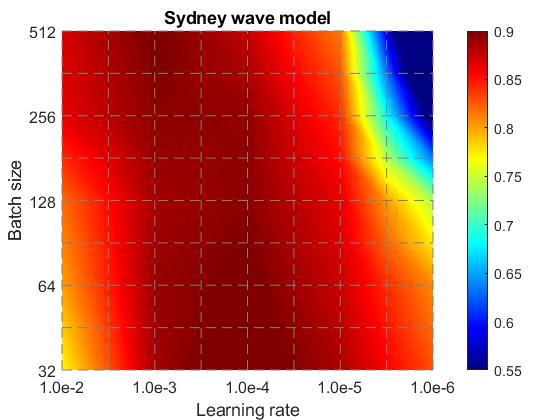}}
 \subfloat[]{
 \includegraphics[clip,width=0.25\columnwidth]{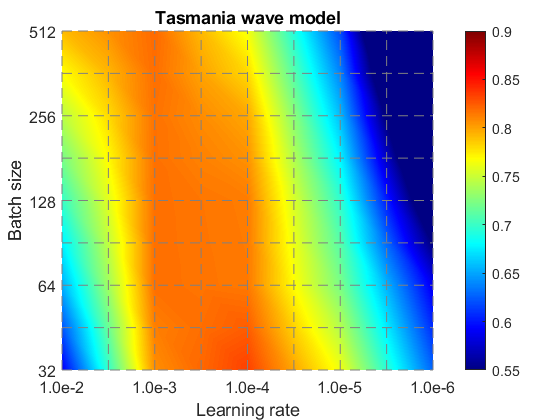}}
 \caption{ CNN-BiLSTM-SA hyper-parameters tuning using grid search for four wave models, a) Adelaide, b) Perth, c) Sydney, d) Tasmania sea sites.   }
\label{fig:hyper_tuning_gridesarch}
\end{figure}

\begin{figure}[h!]
\centering
\subfloat[]{
 \includegraphics[clip,width=0.5\columnwidth]{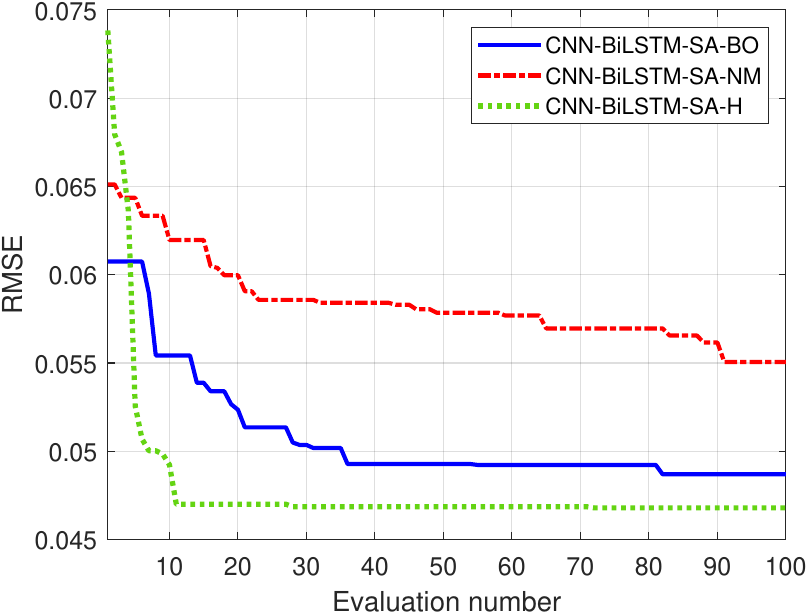}}
 
 \caption{ Convergence rate comparison for CNN-BiLSTM-SA hyper-parameters tuning based on the Tasmania wave model.   }
\label{fig:hyper_tuning_EA}
\end{figure}
\begin{figure}[h!]
\centering
\subfloat[]{
 \includegraphics[clip,width=0.25\columnwidth]{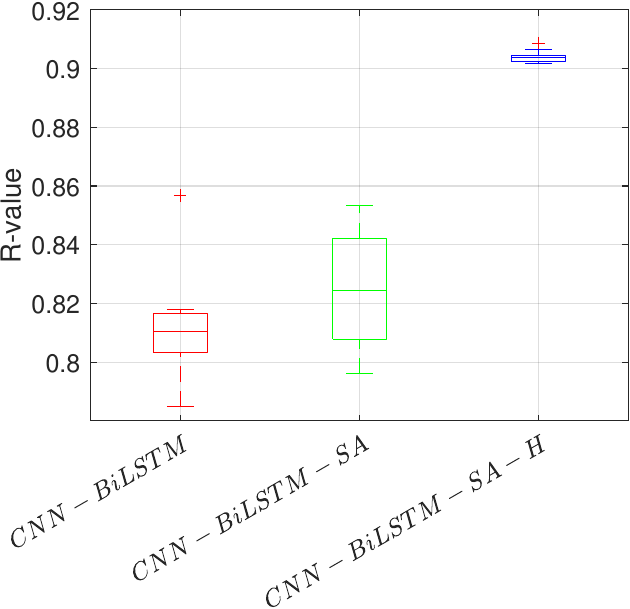}}
 \subfloat[]{
 \includegraphics[clip,width=0.25\columnwidth]{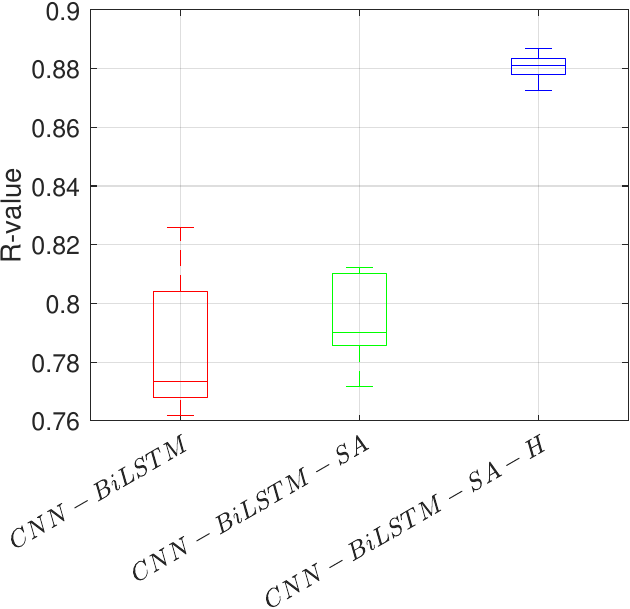}}
 \subfloat[]{
 \includegraphics[clip,width=0.25\columnwidth]{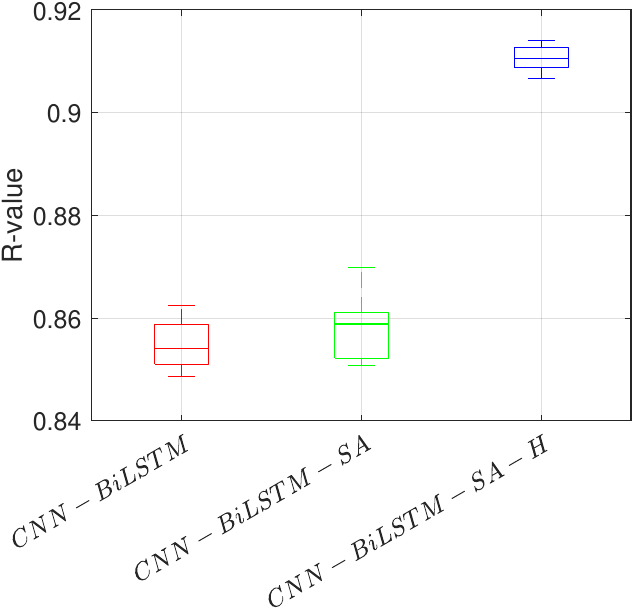}}
 \subfloat[]{
 \includegraphics[clip,width=0.25\columnwidth]{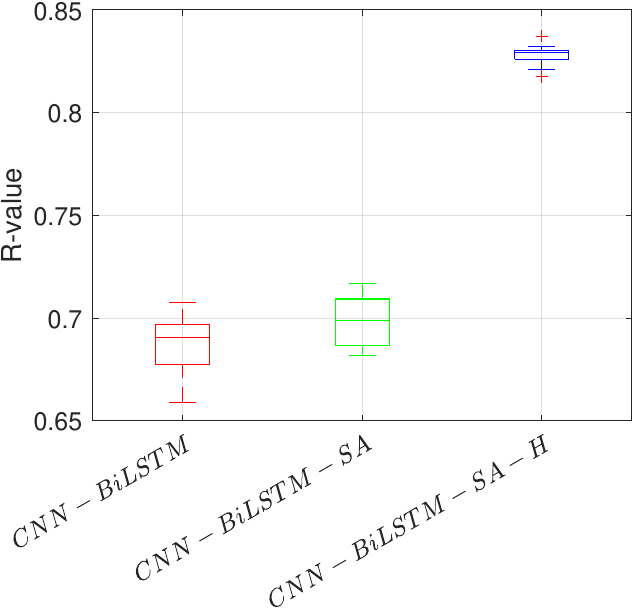}}\\
 \subfloat[]{
 \includegraphics[clip,width=0.24\columnwidth]{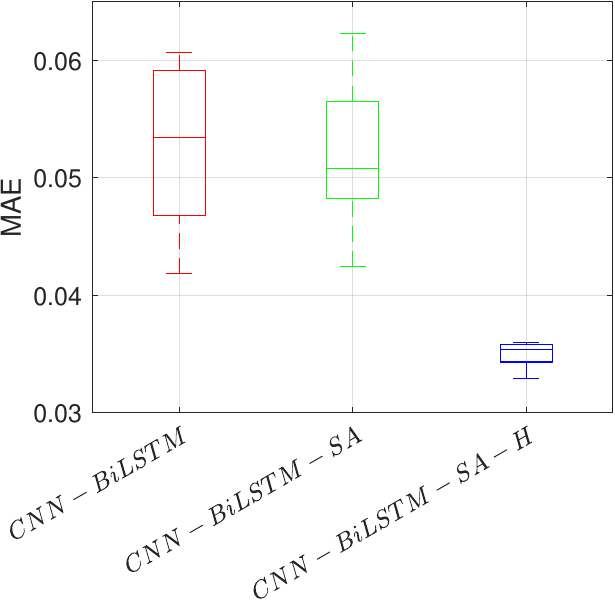}}
 \subfloat[]{
 \includegraphics[clip,width=0.25\columnwidth]{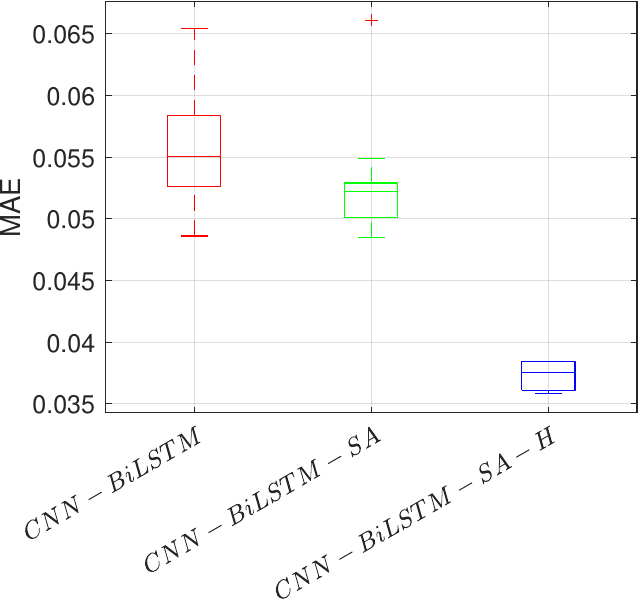}}
 \subfloat[]{
 \includegraphics[clip,width=0.25\columnwidth]{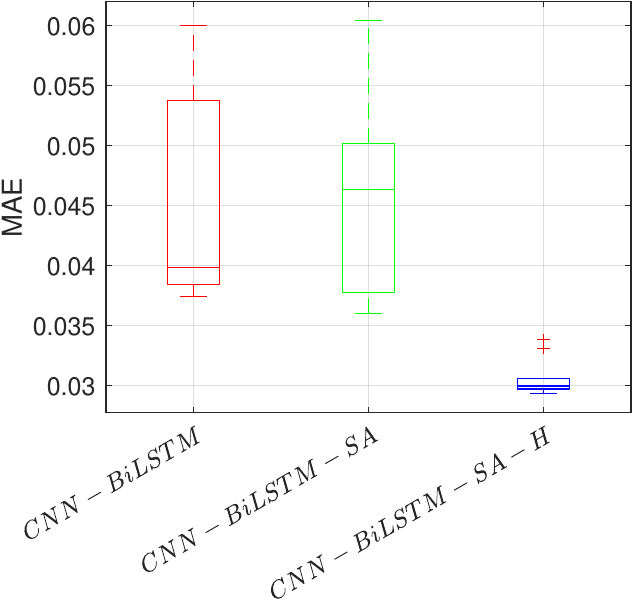}}
 \subfloat[]{
 \includegraphics[clip,width=0.25\columnwidth]{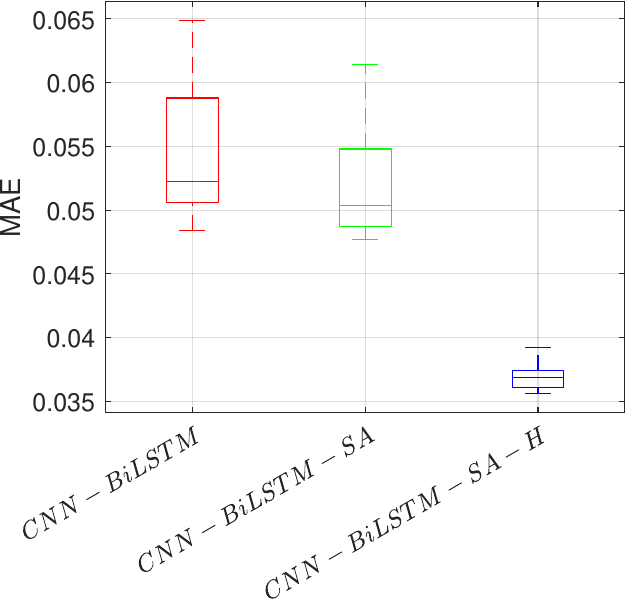}}
 \caption{ A comparison between CNN-BiLSTM, CNN-BiLSTM combined with self-attention, and hybrid of CNN-BiLSTM-SA with an evolutionary hyper-parameter tuner for four wave models, (a, e) Adelaide, (b, f) Perth, (c, g) Sydney, (d, h) Tasmania sea sites.   }
\label{fig:prediction_final}
\end{figure}
Traditional recurrent neural networks (RNNs) like LSTM and GRU encounter difficulties capturing long-range dependencies within sequential data. However, including self-attention layers in hybrid models proves advantageous in addressing this limitation. Self-attention allows each position in the input sequence to attend to all other positions, enabling the hybrid model to capture long-distance relationships and dependencies effectively. This capability enhances the model's contextual understanding and overall performance. Table~\ref{table:results_CNN_Bilstm_SA_H} presents the prediction results of the combined CNN-BiLSTM model with self-attention (SA). A comparison between CNN-BiLSTM-SA and CNN-BiLSTM reveals a notable improvement attributed to the benefits of the self-attention mechanism, resulting in performance gains of  1.72\%, 1.02\%, 1.45\%, and 0.47\% in the respective locations of Adelaide,  Perth, Tasmania, and Sydney. These results further demonstrate the efficacy of self-attention layers in enhancing the performance of hybrid models.

In order to further enhance the performance of the hybrid model  (CNN-BiLSTM-SA), we introduced a fast and adaptive hyper-parameter optimiser named CNN-BiLSTM-SA-H. The outcomes of this modified hybrid model are presented in Table~\ref{table:results_CNN_Bilstm_SA_H}. Containing the hyper-parameter optimiser had a significant impact, resulting in substantial improvements in power prediction accuracy. Specifically, the accuracy rates reached 91\%, 90.4\%, 88\%, and 82.8\% in Sydney, Adelaide,  Perth, and Tasmania, respectively. These results highlight the efficacy of the CNN-BiLSTM-SA-H model in optimising hyper-parameters and further enhancing the accuracy of power predictions in wave energy applications.

Figure~\ref{fig:prediction_models_all} presents a distribution summary of 12 deep learning models, showcasing key statistical measures such as the median, quartiles, and outliers (indicated by red crosses). The accuracy metric ($R^2$) is used to evaluate the wave farm power predictions across four sea sites. This plot enables comparison and visualisation of the predictions' spread and central tendency.

Among the evaluated models, the proposed hybrid model,  CNN-BiLSTM-SA-H, exhibited superior performance compared to standard, stacked, and other hybrid models. Remarkably, the combination with Bi-LSTM yielded more accurate power predictions among the hybrid CNN models. Additionally, CNNs exhibited intriguingly better performance than other  Vanilla models (LSTM, BiLSTM, and GRU). This can be attributed to specific characteristics of CNNs, including their ability to extract local features, share parameters efficiently, exhibit translation invariance, and learn hierarchical representations.

Figure~\ref{fig:prediction_mae_all} presents the statistical prediction results, further emphasising the considerable improvement achieved by the proposed hybrid model across all four wave scenarios, as measured by the mean absolute error (MAE). These findings have significant practical implications for wave energy prediction and deep learning model development, enhancing the accuracy and reliability of wave farm power predictions.

\subsection{Hyper-parameters tuning results}
In Section 3.8, we introduced an efficient and fast hyper-parameter optimiser called the Evolutionary Grid Search (EGS) to fine-tune the hyper-parameters of the hybrid model (CNN-BiLSTM-SA). To gain insights into the impact of the learning rate and batch size on the power prediction performance, we conducted a grid search, and the results are depicted in Figure~\ref{fig:hyper_tuning_gridesarch}. The best-performing hyper-parameter combinations are highlighted in dark red. It is observed that the optimal range for the learning rate lies between $1.0e-3$ and $1.0e-4$, while a batch size greater than 64 yields the best results across all wave farms.

Figure~\ref{fig:hyper_tuning_EA} provides a comparison of the convergence rates among three different hyper-parameter tuners and underscores the superior exploration capability of the proposed method (CNN-BiLSTM-SA-H) in identifying the optimal configurations within the initial evaluation phase.
As summarised in Table~\ref{table:hyper-prameters}, the best-found hyperparameters encompass a range of settings crucial to achieving high performance in predicting the total wave farm power output. CNF and NHU denote the number of filters and hidden units in the convolutional and BiLSTM layers,  respectively. PDO represents the dropout probability for the dropout layers. BS and LR indicate the optimal batch size and learning rate coefficient, respectively. Additionally, L2Reg denotes the adoption of  L2 regularisation, while AT represents the Attention rate.
The comprehensive range of hyper-parameter settings showcased in Table~\ref{table:hyper-prameters} emphasises the significance of thorough hyper-parameter tuning investigations. By identifying the most effective hyper-parameter configurations, we can maximise the model's predictive performance in estimating the total wave farm power output.

To gain a comprehensive understanding of the proposed hybrid model's evolution process and the influence of the self-attention strategy and hyper-parameter optimisation, Figure~\ref{fig:prediction_final} is generated. The boxplots provide valuable insights into the performance improvements achieved by the proposed hyper-parameter optimiser. The results demonstrate its significant role in enhancing power prediction accuracy in the Adelaide, Perth, Sydney, and Tasmania sea sites by approximately 9.8\%, 12.8\%, 5.8\%, and 18.6\%, respectively. These achievements emphasise the hyper-parameter optimisation approach's effectiveness in enhancing the hybrid model's predictive capabilities.

\section{Conclusion}
\label{sec:conclusion}
In this article, we introduce the innovative Self-attention Convolutional Bi-LSTM Technique as a robust strategy for forecasting power generation in wave farms. Through the utilisation of the strengths of Convolutional Neural Networks (CNNs), Bidirectional Long Short-Term Memory networks (Bi-LSTMs), and self-attention combined with a fast and effective hyper-parameters tuning method, our methodology adeptly captures the intricate spatial and temporal relationships present in wave farm datasets, resulting in remarkably precise predictions of power output. The flexibility of the self-attention mechanism empowers the model to dynamically allocate optimal weights to various temporal characteristics, thereby enhancing the accuracy of predictions. To assess the performance and generalisation ability of the proposed model, we tested it using four different wave farms on the Australian coasts, including Adelaide, Perth, Sydney, and Tasmania. 

 Our empirical findings, which are grounded in authentic wave farm data, clearly showcase the supremacy of the proposed predictive technique when contrasted with other cutting-edge models. The integration of this sophisticated approach holds significant promise for optimising energy distribution and streamlining the seamless assimilation of wave farms into existing electricity networks. Subsequent studies could explore adapting this technique to diverse renewable energy sectors and explore additional innovations to enhance predictions' precision. Overall, this research contributes to the advancement of wave energy technologies and provides valuable insights for commercialising and integrating wave power into the renewable energy landscape.

As a result, this assignment can be extended to improve the performance of renewable energy systems by integrating transfer learning and hyper-parameters adjustment for hybrid models. The pre-trained model can accept all types of wave conditions thanks to transfer learning. As a result, models may be created by collecting the acquired knowledge while addressing the issue and applying the many wave scenarios that are related to the exact same issue.

\begin{singlespace} 
\bibliographystyle{unsrt}
\bibliography{cas-refs}
\end{singlespace}





\end{document}